\documentclass[10pt,conference]{IEEEtran}
\usepackage{cite}
\usepackage{amsmath,amssymb,amsfonts}
\usepackage{algorithmic}
\usepackage{graphicx}
\usepackage{textcomp}
\usepackage{xcolor}
\usepackage[hyphens]{url}
\usepackage{fancyhdr}
\usepackage{hyperref}
\usepackage{tikz}
\usepackage{listings}
\usepackage{multirow}
\usepackage{subcaption}
\usepackage[flushleft]{threeparttable}
\usepackage[font=small,labelfont=bf]{caption}
\usepackage{threeparttable}
\usepackage{dblfloatfix}
\usepackage{amssymb}
\usepackage{pifont}
\usepackage{amsmath}
\usepackage{booktabs}
\usepackage[flushleft]{threeparttable}
\usepackage{pifont}
\usepackage{array}
\usepackage{tikz}
\usepackage{todonotes}
\usepackage{balance}

\newcommand{\NAME}{REASON\space}
\newcommand{\NAMEnospace}{REASON}

\newcommand*\circled[1]{\tikz[baseline=(char.base)]{
            \node[shape=circle,fill,inner sep=0.2pt] (char) {\textcolor{white}{#1}};}}

\newcommand{\hpcayear}{2026}

\newcommand{\hpcasubmissionnumber}{996}
\title{\fontsize{23.8}{30.0}\selectfont{REASON: Accelerating Probabilistic Logical Reasoning for Scalable Neuro-Symbolic Intelligence}}

\def\hpcacameraready{} 

\newcommand\hpcaauthors{Zishen Wan$^{\dagger}$, Che-Kai Liu, Jiayi Qian, Hanchen Yang, Arijit Raychowdhury, Tushar Krishna}
\newcommand\hpcaaffiliation{\emph{School of Electrical and Computer Engineering, Georgia Institute of Technology, Atlanta, GA}}
\newcommand\hpcaemail{\{zishenwan, che-kai, jiayiqian, hanchen\}@gatech.edu, \{arijit.raychowdhury, tushar\}@ece.gatech.edu}

\author{
  \ifdefined\hpcacameraready
    \IEEEauthorblockN{\hpcaauthors{}}
      \IEEEauthorblockA{
        \hpcaaffiliation{} \\
        \hpcaemail{}
      }
  \else
    \IEEEauthorblockN{\normalsize{HPCA \hpcayear{} Submission
      \textbf{\#\hpcasubmissionnumber{}}} \\
      \IEEEauthorblockA{
        Confidential Draft \\
        Do NOT Distribute!! 
      }
    }
  \fi 
}

\fancypagestyle{camerareadyfirstpage}{%
  \fancyhead{}
  
  \fancyhead[C]{
    \ifdefined\aeopen
    \parbox[][12mm][t]{13.5cm}{\hpcayear{} IEEE International Symposium on High-Performance Computer Architecture (HPCA)}    
    \else
      \ifdefined\aereviewed
      \parbox[][12mm][t]{13.5cm}{\hpcayear{} IEEE International Symposium on High-Performance Computer Architecture (HPCA)}
      \else
      \ifdefined\aereproduced
      \parbox[][12mm][t]{13.5cm}{\hpcayear{} IEEE International Symposium on High-Performance Computer Architecture (HPCA)}
      \else
      \parbox[][0mm][t]{13.5cm}{\hpcayear{} IEEE International Symposium on High-Performance Computer Architecture (HPCA)}
    \fi 
    \fi 
    \fi 
    \ifdefined\aeopen 
      \includegraphics[width=12mm,height=12mm]{ae-badges/open-research-objects.pdf}
    \fi 
    \ifdefined\aereviewed
      \includegraphics[width=12mm,height=12mm]{ae-badges/research-objects-reviewed.pdf}
    \fi 
    \ifdefined\aereproduced
      \includegraphics[width=12mm,height=12mm]{ae-badges/results-reproduced.pdf}
    \fi
  }
  \fancyfoot[C]{}
}
\begin{document}

\maketitle

\ifdefined\hpcacameraready 
  \thispagestyle{camerareadyfirstpage}
  \pagestyle{empty}
\else
  \thispagestyle{plain}
  \pagestyle{plain}
\fi

\newcommand{\hpcaheight}{0mm}
\ifdefined\eaopen
\renewcommand{\hpcaheight}{12mm}
\fi


\begin{abstract}
Neuro-symbolic AI systems integrate neural perception with symbolic and probabilistic reasoning to enable data-efficient, interpretable, and robust intelligence beyond purely neural models. Although this compositional paradigm has shown superior performance in domains such as mathematical reasoning, planning, and verification, its deployment remains challenging due to severe inefficiencies in symbolic and probabilistic inference. Through systematic analysis of representative neuro-symbolic workloads, we identify probabilistic logical reasoning as the inefficiency bottleneck, characterized by irregular control flow, low arithmetic intensity, uncoalesced memory accesses, and poor hardware utilization on CPUs and GPUs.

This paper presents REASON, an integrated acceleration framework for probabilistic logical reasoning in neuro-symbolic AI. At the algorithm level, REASON introduces a unified directed acyclic graph representation that captures common structure across symbolic and probabilistic models, coupled with adaptive pruning and regularization. At the architecture level, REASON features a reconfigurable, tree-based processing fabric optimized for irregular traversal, symbolic deduction, and probabilistic aggregation. At the system level, REASON is tightly integrated with GPU streaming multiprocessors through a programmable interface and multi-level pipeline that efficiently orchestrates neural, symbolic, and probabilistic execution.
Evaluated across six neuro-symbolic workloads, REASON achieves 12-50$\times$ speedup and 310-681$\times$ energy efficiency over desktop and edge GPUs under TSMC 28 nm node. REASON enables real-time probabilistic logical reasoning, completing end-to-end tasks in 0.8 s with 6~mm$^\text{2}$ area and 2.12~W power, demonstrating that targeted acceleration of probabilistic logical reasoning is critical for practical and scalable neuro-symbolic AI and positioning REASON as a foundational system architecture for next-generation cognitive intelligence.

\end{abstract}
\vspace{-2pt}
\section{Introduction}
\label{sec:intro}
Large Language Models (LLMs) have demonstrated remarkable capabilities in natural language understanding, image recognition, and complex pattern learning from vast datasets~\cite{kuang2025natural,naeem2023i2mvformer,mirchandani2023large,ibrahim2024special}. However, despite their success, LLMs often struggle with factual accuracy, hallucinations, multi-step reasoning, and interpretability~\cite{mahaut2024factual,sriramanan2024llm,aksitov2023rest,singh2024rethinking}. These limitations have spurred the development of \emph{compositional AI systems}, which integrate neural with symbolic and probabilistic reasoning to create robust, transparent, and intelligent cognitive systems.\footnotetext{$^{\dagger}$Corresponding author}

One promising compositional paradigm is neuro-symbolic AI, which integrates neural, symbolic, and probabilistic components into a unified cognitive architecture~\cite{shengyuan2023differentiable,ahmed2022semantic,wan2024towards_mlsys,du2025cross,wan2025efficient}. In this system, the \emph{neural} module captures the statistical, pattern-matching behavior of learned models, performing rapid function approximation and token prediction for intuitive perception and feature extraction. The \emph{symbolic} and \emph{probabilistic} modules perform explicit, verifiable reasoning that is structured, interpretable, and robust under uncertainty, managing logic-based reasoning and probabilistic updates. This paradigm integrates intuitive generalization and deliberate reasoning.


Neuro-symbolic AI has demonstrated superior abstract deduction, complex question answering, mathematical reasoning, logical reasoning, and cognitive robotics~\cite{li2025proving,trinh2024solving,romera2024mathematical,zhang2021abstract,hersche2023neuro,mao2019neuro,mei2022falcon,wan2025reca}. Its ability to learn efficiently from fewer data points, produce transparent and verifiable outputs, and robustly handle uncertainty and ambiguity makes it particularly advantageous compared to purely neural approaches. For example, recently Meta's LIPS~\cite{li2025proving} and Google's AlphaGeometry~\cite{trinh2024solving} leverage compositional neuro-symbolic approaches to solve complex math problems and achieve a level of human Olympiad gold medalists. R$^2$-Guard~\cite{kang2024r} leverages LLM and probabilistic models to improve robust reasoning capability and resilience against jailbreaks. They represent a paradigm shift for AI that requires robust, verifiable, and explainable reasoning. 

Despite impressive algorithmic advances in neuro-symbolic AI -- often demonstrated on large-scale distributed GPU clusters -- efficient deployment at the edge remains a fundamental challenge. Neuro-symbolic agents, particularly in robotics, planning, interactive cognition, and verification, require \emph{real-time} logical inference to interact effectively with physical environments and multi-agent systems. For example, Ctrl-G, a text-infilling neuro-symbolic agent~\cite{zhang2024adaptable}, must execute hundreds of reasoning steps per second to remain responsive, yet current implementations take over 5 minutes on a desktop GPU to complete a single task. This latency gap makes practical deployment of neuro-symbolic AI systems challenging.

To understand the root causes of this inefficiency, we systematically analyze a diverse set of neuro-symbolic workloads and uncover several system- and architecture-level challenges. Symbolic and probabilistic kernels frequently dominate end-to-end runtime and exhibit highly irregular execution characteristics, including heterogeneous compute patterns and memory-bound behavior with low ALU utilization. These kernels suffer from limited exploitable parallelism and irregular, uncoalesced memory accesses, leading to poor performance and efficiency on CPU and GPU architectures.

To address these challenges, we develop an integrated acceleration framework, \NAMEnospace, which to the best of our knowledge, is the first to accelerate probabilistic logical reasoning-based neuro-symbolic AI systems. \NAME is designed to close the efficiency gap of compositional AI by jointly optimizing algorithms, architecture, and system integration for the irregular and heterogeneous workloads inherent to neuro-symbolic reasoning.

At the algorithm level, \NAME introduces a unified directed acyclic graph (DAG) representation that captures shared computational structure across symbolic and probabilistic kernels. An adaptive pruning and regularization technique further reduces model size and computational complexity while preserving task accuracy.
At the architecture level, \NAME features a flexible design optimized for various irregular symbolic and probabilistic computations, leveraging the unified DAG representation. The architecture comprises reconfigurable tree-based processing elements (PEs), compiler-driven workload mapping, and memory layout to enable highly parallel and energy-efficient symbolic and probabilistic computation.
At the system level, \NAME is tightly integrated with GPU streaming multiprocessors (SMs), forming a heterogeneous system with a programmable interface and multi-level execution pipeline that efficiently orchestrates neural, symbolic, and probabilistic kernels while maintaining high hardware utilization and scalability as neuro-symbolic models evolve. 
Notably, unlike conventional tree-like computing arrays optimized primarily for neural workloads, \NAME provides reconfigurable support for neural, symbolic, and probabilistic kernels within a unified execution fabric, enabling efficient and scalable neuro-symbolic AI systems.

This paper, therefore, makes the following contributions:
\begin{itemize}
\item We conduct a systematic workload characterization of representative logical- and probabilistic-reasoning-based neuro-symbolic AI models, identifying key performance bottlenecks and architectural optimization opportunities (Sec.~\ref{sec:background}, Sec.~\ref{sec:profiling}).

\item We propose \NAMEnospace, an integrated co-design framework, to efficiently accelerate probabilistic logical reasoning in neuro-symbolic AI, enabling practical and scalable deployment of compositional intelligence (Fig.~\ref{fig:leading}).

\item \NAME introduces cross-layer innovations spanning (i) a unified DAG representation with adaptive pruning at the algorithm level (Sec.~\ref{sec:algo_opt}), (ii) a reconfigurable symbolic/probabilistic architecture and compiler-driven dataflow and mapping at the hardware level (Sec.~\ref{sec:hardware}), and (iii) a programmable system interface with a multi-level execution pipeline at the system level (Sec.~\ref{sec:system}) to improve neuro-symbolic efficiency.

\item Evaluated across cognitive tasks, \NAME enables flexible support for symbolic and probabilistic operations, achieving 12-50$\times$ speedup and 310-681$\times$ energy efficiency compared to desktop and edge GPUs. \NAME enables fast and efficient logical and probabilistic reasoning in 0.8 s per task with 6~mm$^2$ area and 2.12~W power consumption. (Sec.~\ref{sec:eval}).
\end{itemize}

\section{Neuro-Symbolic AI Systems}
\label{sec:background}

This section presents the preliminaries of neuro-symbolic AI with its algorithm flow (Sec.~\ref{subsec:background_workflow}), scaling performance analysis (Sec.~\ref{subsec:scaling_analysis}), and key computational primitives (Sec.~\ref{subsec:background_ops}).


\begin{figure}[t!]
\centering
\includegraphics[width=\columnwidth]{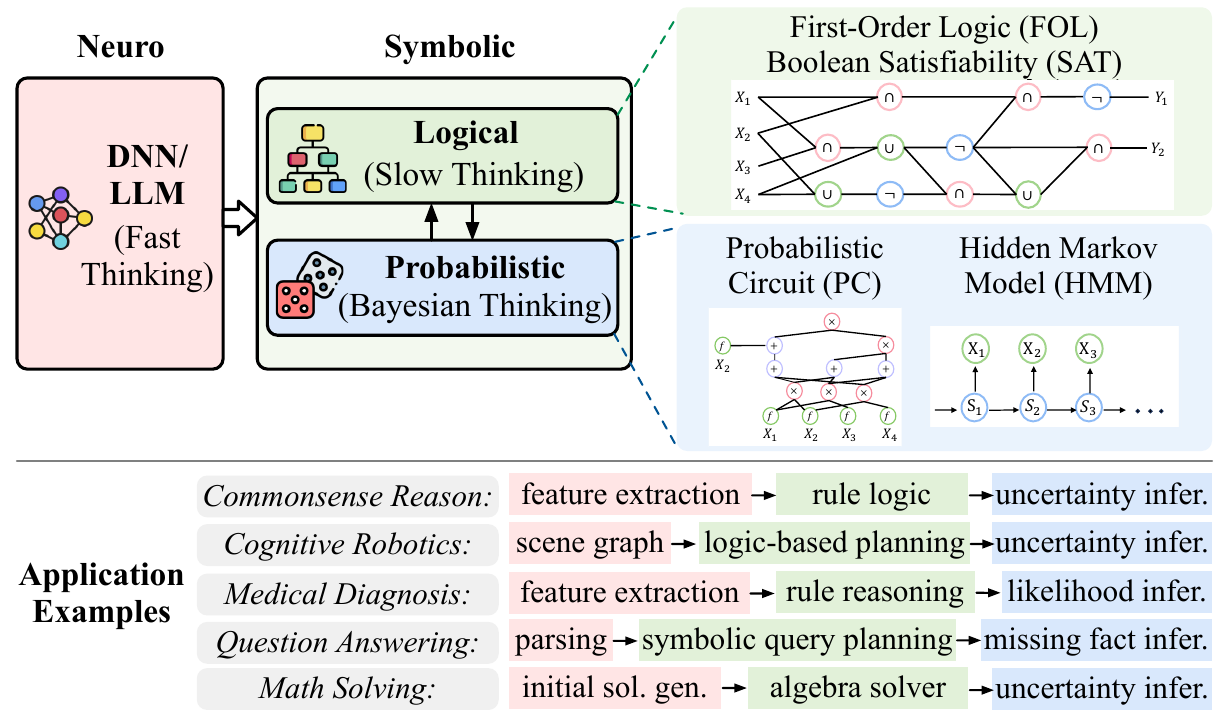}
\vspace{-0.1in}
        \caption{\textbf{Neuro-symbolic algorithmic flow and operations.} The neural module serves as a perceptual and intuitive engine for representation learning, while the symbolic module performs structured logical reasoning with probabilistic inference. This compositional pipeline enables complex cognitive tasks across diverse scenarios.}
    \label{fig:algo_ops}
        \vspace{-10pt}
\end{figure}

\subsection{Neuro-Symbolic Cognitive Intelligence}
\label{subsec:background_workflow}
\label{subsec:background_challenge}


LLMs and DNNs excel at natural language understanding and image recognition. However, they remain prone to factual errors, hallucinations, challenges in complex multi-step reasoning, and vulnerability to out-of-distribution or adversarial inputs. Their black-box nature also impedes interpretability and formal verification, undermining trust in safety-critical domains. These limitations motivate the development of \emph{compositional} systems that integrate neural models with symbolic and probabilistic reasoning to achieve greater robustness, transparency, and intelligence.

Neuro-symbolic AI represents a paradigm shift toward more integrated and trustworthy intelligence by combining neural, symbolic, and probabilistic techniques. This hybrid approach has shown superior performance in abstract deduction~\cite{zhang2021abstract,hersche2023neuro}, complex question answering~\cite{mao2019neuro,mei2022falcon}, and logical reasoning~\cite{trinh2024solving,romera2024mathematical}. By learning from limited data and producing transparent, verifiable outputs, neuro-symbolic systems provide a foundation for cognitive intelligence. Fig.~\ref{fig:algo_ops} presents a unified neuro-symbolic pipeline, illustrating how its components collaborate to solve complex tasks.

\begin{figure*}[t!]
\centering\includegraphics[width=\linewidth]{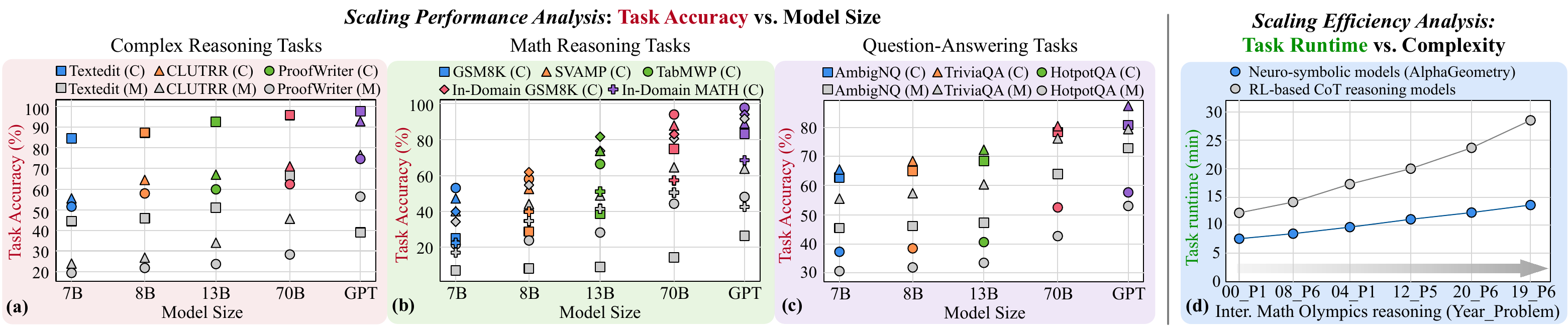}
\vspace{-0.15in}
        \caption{\textbf{Scaling performance and efficiency.} \textbf{(a)-(c)} Task accuracy of compositional LLM-symbolic models (C) and monolithic LLMs (M - shown in \textcolor{gray}{gray}) across model sizes on complex reasoning, mathematical reasoning, and question-answering tasks. \textbf{(d)} Runtime efficiency comparison between LLM-symbolic models and RL-based CoT models on mathematical reasoning tasks~\cite{wan2026compositional}.}
        \label{fig:scaling_analysis}
        \vspace{-10pt}
\end{figure*}

\textbf{Neural module.} The neural module serves as the perceptual and intuitive engine, typically DNN or LLM, excelling at processing high-dimensional sensory inputs (e.g., images, audio, text) and converting them into feature representations. It handles perception, feature extraction, and associative learning, providing the abstractions needed for higher-level cognition.


\textbf{Symbolic module.} The symbolic module is the logical core operating on neural abstractions and includes symbolic and probabilistic operations. \emph{Logical} components apply formal logic, rules, and ontologies for structured reasoning and planning, enabling logically sound solutions. \emph{Probabilistic} components manage uncertainty by representing knowledge probabilistically, supporting belief updates and decision-making under ambiguity, reflecting a nuanced reasoning model.  
 
Together, these modules form a complementary reasoning hierarchy. Neural module captures statistical, pattern-matching behavior of learned models, producing rapid but non-verifiable outputs (\emph{Fast Thinking}), while symbolic modules perform explicit, verifiable reasoning that is structured and reliable (\emph{Slow Thinking}). The probabilistic module complements both and enables robust planning under ambiguity (\emph{Bayesian Thinking}). This framework integrates intuitive generalization with deliberate reasoning.

\subsection{Scaling Performance Analysis of Neuro-Symbolic Systems}
\label{subsec:scaling_analysis}

\textbf{Scaling performance analysis.} Neuro-symbolic AI systems exhibit superior reasoning capability and scaling behavior compared to monolithic LLMs on complex tasks. We compare representative neuro-symbolic systems against monolithic LLMs across complex reasoning, mathematical reasoning, and question-answering benchmarks (Fig.~\ref{fig:scaling_analysis}(a)-(c)).  
The results reveal two advantages.  
\underline{\emph{First}}, \emph{higher accuracy}: compositional neuro-symbolic models consistently outperform monolithic LLMs of comparable size.  
\underline{\emph{Second}}, \emph{improved scaling efficiency}: smaller neuro-symbolic models are sufficient to match or exceed the performance of significantly larger closed-source LLMs.  
Together, these results highlight the potential scaling limitations of monolithic LLMs and the efficiency benefits of compositional neuro-symbolic reasoning.

\textbf{Comparison with RL-based reasoning models.} Beyond monolithic LLMs, recent advancements in reinforcement learning (RL) and chain-of-thought (CoT) prompting improve LLM reasoning accuracy but incur significant computational and scalability overheads (Fig.~\ref{fig:scaling_analysis}(d)). 
\emph{\underline{First}, computational cost}: RL-based reasoning often requires hundreds to thousands of LLM queries per decision step, resulting in prohibitively high inference latency and energy consumption. 
\emph{\underline{Second}, scalability}: task-specific fine-tuning constrains generality, whereas neuro-symbolic models use symbolic and probabilistic reasoning modules or tools without retraining.  
Fig.~\ref{fig:scaling_analysis}(d) reveals that neuro-symbolic models AlphaGeometry~\cite{trinh2024solving} achieve over $2\times$ efficiency gains and superior data efficiency compared to CoT-based LLMs on mathematical reasoning tasks.

\subsection{Computational Primitives in Neuro-Symbolic AI}
\label{subsec:background_ops}

We identify the core computational primitives that are commonly used in neuro-symbolic AI systems (Fig.~\ref{fig:algo_ops}). While neural modules rely on DNNs or LLMs for perception and representation learning, the symbolic and probabilistic components implement structured reasoning. In particular, \emph{logical reasoning} is typically realized through First-Order Logic (FOL) and Boolean Satisfiability (SAT), \emph{probabilistic reasoning} through Probabilistic Circuits (PCs), and \emph{sequential reasoning} through Hidden Markov Models (HMMs). Together, these primitives form the algorithmic foundation of neuro-symbolic systems that integrate learning, logic, and uncertainty-aware inference.

\textbf{First-Order Logic (FOL) and Boolean Satisfiability (SAT).}  
FOL provides a formal symbolic language for representing structured knowledge using predicates, functions, constants, variables and quantifiers ($\forall,\exists$), combined with logical connectives. For instance, the statement ``every student has a mentor'' can be expressed as 
$\forall x\bigl(\mathrm{Student}(x)\to\exists y\,(\mathrm{Mentor}(y)\wedge\mathrm{hasMentor}(x,y))\bigr)$, where predicates encode properties and relations over domain elements. FOL semantics map symbols to domain objects and relations, enabling precise and interpretable logical reasoning. SAT operates over propositional logic and asks whether a conjunctive normal form (CNF) formula $\varphi = \bigwedge_{i=1}^m \Bigl(\bigvee_{j=1}^{k_i} l_{ij}\Bigr)$ admits a satisfying assignment, where each literal $l_{ij}$ is a Boolean variable or its negation. Modern SAT solvers extend the DPLL algorithm with conflict-driven clause learning (CDCL), incorporating non-chronological backtracking and clause learning to improve scalability~\cite{marques2021conflict,lo2025sat}. Cube-and-conquer further parallelizes search by splitting into ``cube'' DPLL subproblems and concurrent CDCL ``conquer'' solvers~\cite{heule2011cube,van2012concurrent}.
Together, FOL’s expressive representations and SAT’s solving mechanisms form the logic backbone of neuro-symbolic systems, enabling exact logical inference alongside neural learning.


\begin{table*}[t!]
\centering
\scriptsize
\renewcommand*{\arraystretch}{1.2}
\setlength\tabcolsep{5pt}
\resizebox{\linewidth}{!}{%
\begin{tabular}{cc|c|c|c|c|c|c}
\hline
\multicolumn{2}{c|}{\textbf{\begin{tabular}[c]{@{}c@{}}Representative Neuro-\\Symbolic Workloads\end{tabular}}} &
\textbf{\begin{tabular}[c]{@{}c@{}}AlphaGeometry~\cite{trinh2024solving}\end{tabular}} &
\textbf{\begin{tabular}[c]{@{}c@{}}R$^2$-Guard~\cite{kang2024r}\end{tabular}} &
\textbf{\begin{tabular}[c]{@{}c@{}}GeLaTo~\cite{zhang2023tractable}\end{tabular}} &
\textbf{\begin{tabular}[c]{@{}c@{}}Ctrl-G~\cite{zhang2024adaptable}\end{tabular}} &
\textbf{\begin{tabular}[c]{@{}c@{}}NeuroPC~\cite{chen2025neural}\end{tabular}} &
\textbf{\begin{tabular}[c]{@{}c@{}}LINC~\cite{olausson2023linc}\end{tabular}}
\\ \hline

\multicolumn{1}{c|}{\multirow{4}{*}{\textbf{\begin{tabular}[c]{@{}c@{}}Deployment\\ Scenario\end{tabular}}}} &
\textbf{Application} &
\begin{tabular}[c]{@{}c@{}}Math theorem\\ proving \& reasoning\end{tabular} &
\begin{tabular}[c]{@{}c@{}}Unsafety\\ detection\end{tabular} &
\begin{tabular}[c]{@{}c@{}}Constrained\\ text generation\end{tabular} &
\begin{tabular}[c]{@{}c@{}}Interactive text editing,\\ text infilling\end{tabular} &
\begin{tabular}[c]{@{}c@{}}Classification\end{tabular} &
\begin{tabular}[c]{@{}c@{}}Logical reasoning,\\ Deductive reasoning\end{tabular}
\\ \cline{2-8}

\multicolumn{1}{c|}{} &
\textbf{\begin{tabular}[c]{@{}c@{}}Advantage\\ vs. LLM\end{tabular}} &
\begin{tabular}[c]{@{}c@{}}Higher deductive\\ reasoning, higher\\ generalization\end{tabular} &
\begin{tabular}[c]{@{}c@{}}Higher LLM resilience,\\ higher data efficiency,\\ effective adaptability\end{tabular} &
\begin{tabular}[c]{@{}c@{}}Guaranteed constraint\\ satisfaction, higher\\ generalization\end{tabular} &
\begin{tabular}[c]{@{}c@{}}Guaranteed constraints\\ satisfaction, higher\\ generalization\end{tabular} &
\begin{tabular}[c]{@{}c@{}}Enhanced interpretability,\\ theoretical guarantee\end{tabular} &
\begin{tabular}[c]{@{}c@{}}Higher precision,\\ reduced overconfidence,\\ higher scalability\end{tabular}
\\ \hline

\multicolumn{1}{c|}{\multirow{4}{*}{\textbf{\begin{tabular}[c]{@{}c@{}}Computation\\ Pattern\end{tabular}}}} &
\textbf{Neural} &
LLM & LLM & LLM & LLM & DNN & LLM
\\ \cline{2-8}

\multicolumn{1}{c|}{} &
\textbf{Symbolic} &
\begin{tabular}[c]{@{}c@{}}First-order logic,\\ SAT solver,\\ acyclic graph\end{tabular} &
\begin{tabular}[c]{@{}c@{}}First-order logic,\\ probabilistic circuit,\\ Hidden Markov model\end{tabular} &
\begin{tabular}[c]{@{}c@{}}First-order logic,\\ SAT solver,\\ Hidden Markov model\end{tabular} &
\begin{tabular}[c]{@{}c@{}}Hidden Markov model,\\ probabilistic circuits\end{tabular} &
\begin{tabular}[c]{@{}c@{}}First-order logic,\\ probabilistic circuit\end{tabular} &
\begin{tabular}[c]{@{}c@{}}First-order logic,\\ solver\end{tabular}
\\ \hline
\end{tabular}%
}
\caption{\textbf{Representative neuro-symbolic workloads.} Selected neuro-symbolic workloads used in our analysis, spanning diverse application domains, deployment scenarios, and neural-symbolic computation patterns.}
\vspace{-4pt}
\label{tab:selected_workload}
\end{table*}

\begin{figure*}[t!]
\centering\includegraphics[width=2.05\columnwidth]{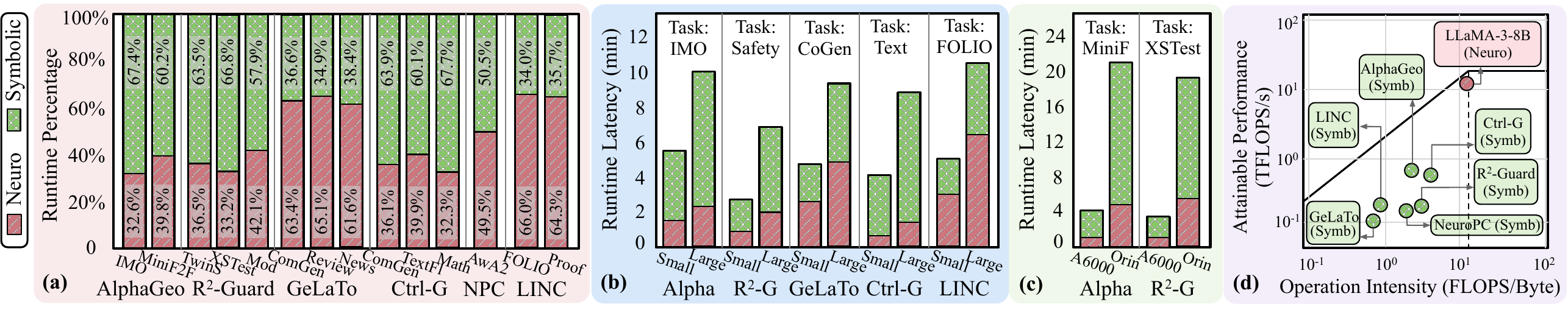}
        \caption{\textbf{End-to-end neuro-symbolic workload characterization.} \textbf{(a)} Benchmark six neuro-symbolic workloads (AlphaGeometry, R$^2$-Guard, GeLaTo, Ctrl-G, NeuroPC, LINC) on CPU+GPU system, showing symbolic and probabilistic may serve as system bottlenecks. \textbf{(b)} Benchmark neuro-symbolic workloads on tasks with different scales, indicating that real-time performance cannot be satisfied and the potential efficiency issues. \textbf{(c)} Benchmark on A6000 and Orin GPU. \textbf{(d)} Roofline analysis, indicating server memory-bound of symbolic and probabilistic kernels.}
        \label{fig:profiling_latency}
        \vspace{-10pt}
\end{figure*}


\textbf{Probabilistic Circuits (PC).}  
PCs represent tractable probabilistic models over variables $\mathbf{X}$ as directed acyclic graphs~\cite{liang2019learning,khosravi2019tractable,liu2024scaling}. Each node $n$ performs a probabilistic computation: leaf nodes specify primitive distributions $f_n(x)$, while interior nodes combine their children $ch(n)$ via  
\begin{equation}
p_n(x) = 
\begin{cases}
f_n(x), & \text{if } n \text{ is a leaf node} \\
\prod_{c \in \mathrm{ch}(n)} p_c(x), & \text{if } n \text{ is a product node} \\
\sum_{c \in \mathrm{ch}(n)} \theta_{n,c} p_c(x), & \text{if } n \text{ is a sum node}
\end{cases}
\label{eq:pc_definition}
\end{equation}
where $\theta_{n,c}$ denotes the non-negative weight associated with child $c$. This recursive structure guarantees exact inference (e.g., marginals, conditionals) in time linear in circuit size.  PCs’ combination of expressiveness and tractable computation makes them an ideal probabilistic backbone for neuro-symbolic systems, where neural modules learn circuit parameters while symbolic engines perform probabilistic reasoning.

\textbf{Hidden Markov Model (HMM).} HMMs are probabilistic model for sequential data~\cite{mor2021systematic}, where a system evolves through hidden states governed by the first-order Markov property: the state at time step $t$ depends only on the state at time step $t-1$. Each hidden state emits observations according to a probabilistic distribution. The joint distribution over sequence of hidden states \(z_{1:T}\) and observations \(x_{1:T}\) is given by
\vspace{-5pt}
\begin{equation}
    p(z_{1:T}, x_{1:T}) = p(z_1) p(x_1\mid z_1) \prod_{t=2}^T p(z_t \mid z_{t-1}) p(x_t \mid z_t)
    \label{eq:hmm}
\end{equation}
where \(p(z_1)\) is the initial state distribution, \(p(z_t\mid z_{t-1})\) the transition probability, and \(p(x_t \mid z_t)\) the emission probability. HMMs naturally support sequential inference tasks such as filtering, smoothing, and decoding, enabling temporal reasoning in neuro-symbolic pipelines.


\section{Neuro-Symbolic Workload Characterization}
\label{sec:profiling}

This section characterizes the system behavior of various neuro-symbolic workloads (Sec.~\ref{subsec:profiling_latency}-\ref{subsec:profiling_nsight}) and provides workload insights for computer architects (Sec.~\ref{subsec:profiling_summary}-\ref{subsec:profiling_opportunities}). 

\textbf{Profiling workloads.} To conduct comprehensive profiling analysis, we select six state-of-the-art representative neuro-symbolic workloads, as listed in Tab.~\ref{tab:selected_workload}, covering a diverse range of applications and underlying computational patterns. 
\label{subsec:profiling_workload}

\textbf{Profiling setup.} We profile and analyze the selected neuro-symbolic models in terms of runtime, memory, and compute operators using cProfile for latency measurement, and NVIDIA Nsight for kernel-level profiling and analysis. 
Experiments are conducted on the system with NVIDIA A6000 GPU, Intel Sapphire Rapids CPUs, and DDR5 DRAM. Our software environment includes PyTorch 2.5 and JAX 0.4.6. We also conduct profiling on Jetson Orin~\cite{orin} for edge scenario deployment. We track control and data flow by analyzing the profiling results in trace view and graph execution format.

\subsection{Compute Latency Analysis}
\label{subsec:profiling_latency}
\label{subsec:profiling_setup}


\begin{figure*}[b]
\centering\includegraphics[width=2.05\columnwidth]{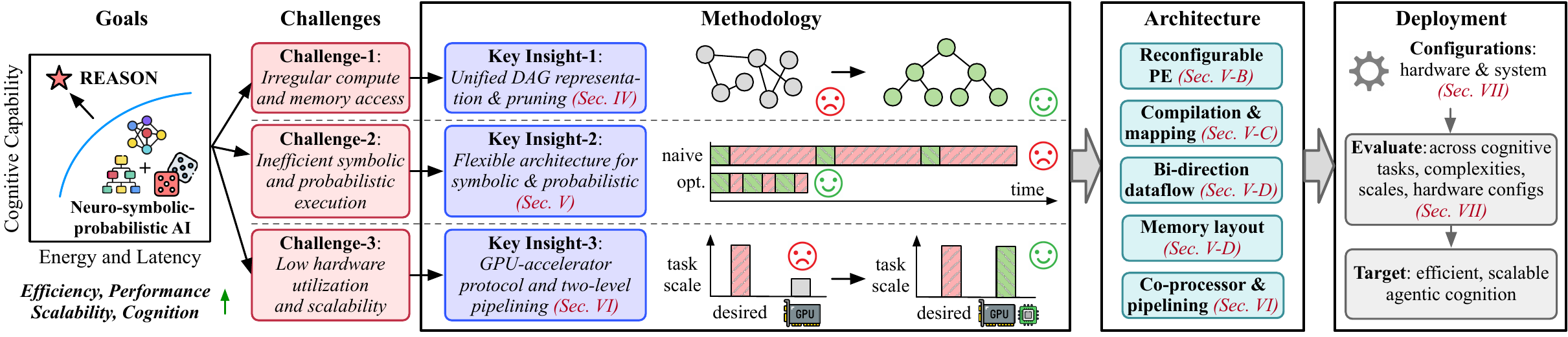}
        \caption{\textbf{\NAME overview.} \textnormal{\NAME is an integrated acceleration framework for probabilistic logical reasoning grounded neuro-symbolic AI with the \underline{goal} to achieve efficient and scalable agentic cognition. \NAME addresses the \underline{challenges} of irregular compute and memory, symbolic and probabilistic latency bottleneck, and hardware underutilization, by proposing \underline{methodologies} including unified DAG representation, reconfigurable PE, efficient dataflow, mapping, scalable architecture, two-level parallelism and programming interface. \NAME is \underline{deployed} across cognitive tasks and consistently demonstrates performance-efficiency improvements for compositional neuro-symbolic systems.}
        }
        \label{fig:leading}
        \vspace{-10pt}
\end{figure*}

\textbf{Latency bottleneck.} We characterize the latency of representative neuro-symbolic workloads (Fig.~\ref{fig:profiling_latency}(a)).  
\emph{Compared to neuro kernels, symbolic and probabilistic kernels are not negligible in latency and may become system bottlenecks}.  
For instance, the neural (symbolic) components account for 36.2\% (63.8\%), 37.3\% (62.7\%), 63.4\% (36.6\%), 36.1\% (63.9\%), 49.5\% (50.5\%), and 65.2\% (34.8\%) of runtime in AlphaGeometry, R$^2$-Guard, GeLaTo, Ctrl-G, NeuroPC, and LINC, respectively.  
Symbolic kernels dominate AlphaGeometry’s runtime, and probabilistic kernels dominate R$^2$-Guard and Ctrl-G’s, due to high irregular memory access, wrap divergence, thread underutilization, and execution parallelism.  
FLOPS and latency measurements further highlight this inefficiency. 
Notably, when using a smaller LLM (LLaMA-7B) for GeLaTo and LINC, overall accuracy remains stable, but the symbolic latency rises to 69.0\% and 65.5\%, respectively.
We observe consistent trends in the Orin NX-based platform (Fig.~\ref{fig:profiling_latency}(c)). Symbolic components count for 63.8\% of AlphaGeometry runtime on A6000 while its FLOPS count for only 19.3\%, indicating inefficient hardware utilization.

\textbf{Latency scalability.} We evaluate runtime across reasoning tasks of varying difficulty and scale (Fig.~\ref{fig:profiling_latency}(b)). We observe that the relative runtime distribution between neural and symbolic components remains consistent of a single workload across task sizes.
Total runtime increases with task complexity and scale. While LLM kernels scale efficiently due to their tensor-based GPU-friendly inference, logical and probabilistic kernels scale poorly due to the exponential growth of search space, making them slower compared to monolithic LLMs.


\subsection{Roofline \& Symbolic Operation \& Inefficiency Analysis}
\label{subsec:profiling_roofline}
\textbf{Memory-bounded operation.} Fig.~\ref{fig:profiling_latency}(d) presents a roofline analysis of GPU memory bandwidth versus compute efficiency. \emph{We observe that the symbolic and probabilistic components are typically memory-bound, limiting performance efficiency}. 
For example, R$^2$-Guard’s probabilistic circuits use sparse, scattered accesses for marginalization, and Ctrl-G’s HMM iteratively reads and writes state probabilities.  
Low compute per element makes these workloads constrained by memory access, underutilizing GPU compute resources.

\label{subsec:profiling_nsight}
\begin{table}[t!]
\scriptsize
\centering
\vspace{5pt}
\caption{\textbf{Hardware inefficiency analysis.} \textnormal{The compute, memory, and communication characteristics of representative neural, symbolic, and probabilistic kernels executed on CPU/GPU platform.}}
\renewcommand*{\arraystretch}{1.05}
\newcolumntype{P}[1]{>{\raggedright\arraybackslash}p{#1}}
\setlength\tabcolsep{2.2pt}
\resizebox{\linewidth}{!}{%
\begin{tabular}{l|cc|cc|cc}
\hline
\multirow{2}{*}{}        & \multicolumn{2}{c|}{\textbf{Neural Kernel}}                         & \multicolumn{2}{c|}{\textbf{Symbolic Kernel}} & \multicolumn{2}{c}{\textbf{Probabilistic Kernel}}                \\ \cline{2-7} 
                         & \multicolumn{1}{c|}{MatMul} & Softmax & \multicolumn{1}{c|}{Sparse MatVec} & Logic & \multicolumn{1}{c|}{Marginal} & Bayesian \\ \hline
\multicolumn{7}{l}{\textit{Compute Efficiency}} \\ \hline
Compute Throughput (\%)  & \multicolumn{1}{c|}{96.8} & 62.2 & \multicolumn{1}{c|}{32.5} & 14.7 & \multicolumn{1}{c|}{35.0} & 31.1 \\ \hline
ALU Utilization (\%)     & \multicolumn{1}{c|}{98.4} & 72.0 & \multicolumn{1}{c|}{43.9} & 29.3 & \multicolumn{1}{c|}{48.5} & 52.8 \\ \hline
\multicolumn{7}{l}{\textit{Memory Behavior}} \\ \hline
L1 Cache Throughput (\%) & \multicolumn{1}{c|}{82.4} & 58.0 & \multicolumn{1}{c|}{27.1} & 20.6 & \multicolumn{1}{c|}{32.4} & 37.1 \\
L2 Cache Throughput (\%) & \multicolumn{1}{c|}{41.7} & 27.6 & \multicolumn{1}{c|}{18.3} & 12.4 & \multicolumn{1}{c|}{24.2} & 27.5 \\ \hline
L1 Cache Hit Rate (\%)   & \multicolumn{1}{c|}{88.5} & 85.0 & \multicolumn{1}{c|}{53.6} & 37.0 & \multicolumn{1}{c|}{42.4} & 40.7 \\
L2 Cache Hit Rate (\%)   & \multicolumn{1}{c|}{73.4} & 66.7 & \multicolumn{1}{c|}{43.9} & 32.7 & \multicolumn{1}{c|}{50.2} & 47.6 \\ \hline
DRAM BW Utilization (\%) & \multicolumn{1}{c|}{39.8} & 28.6 & \multicolumn{1}{c|}{57.4} & 70.3 & \multicolumn{1}{c|}{60.8} & 68.0 \\ \hline
\multicolumn{7}{l}{\textit{Control Divergence and Scheduling}} \\ \hline
Warp Execution Efficiency (\%) & \multicolumn{1}{c|}{96.3} & 94.1 & \multicolumn{1}{c|}{48.8} & 54.0 & \multicolumn{1}{c|}{59.3} & 50.6 \\ \hline
Branch Efficiency (\%) & \multicolumn{1}{c|}{98.0} & 98.7 & \multicolumn{1}{c|}{60.0} & 58.1 & \multicolumn{1}{c|}{63.4} & 66.9 \\ \hline
Eligible Warps/Cycle (\%) & \multicolumn{1}{c|}{7.2} & 7.0 & \multicolumn{1}{c|}{2.4} & 2.1 & \multicolumn{1}{c|}{2.8} & 2.5 \\ \hline
\end{tabular}
}
\label{tab:nsight_profile}
\vspace{-10pt}
\end{table}

\textbf{Hardware inefficiency analysis.}
We leverage Nsight Systems and Nsight Compute~\cite{nsightsystem,nsightcompute} to analyze the computational, memory, and control \emph{irregularity} of neural, symbolic, and probabilistic kernels, as listed in Tab.~\ref{tab:nsight_profile}. We observe that: 
\emph{\underline{First}, compute throughput and ALU utilization}: neural kernels achieve high throughput and ALU utilization, while symbolic/probabilistic kernels have low throughput and idle ALUs.  
\emph{\underline{Second}, memory access and cache utilization}: neural kernels see high L1 cache hit rates; symbolic kernels incur cache misses and stalls, and probabilistic kernels face high memory pressure.  
\emph{\underline{Third}, DRAM bandwidth (BW) utilization and data movement overhead}: neural workloads use on-chip caches with minimal DRAM usage, but symbolic/probabilistic workloads are DRAM-bound with heavy random-access overhead. 

\textbf{Sparsity analysis.} We observe high, heterogeneous, irregular, and data-dependent sparsity across neuro-symbolic workloads. Symbolic and probabilistic kernels are often extremely sparse, exhibiting on average 82\%, 87\%, 75\%, 83\%, 89\%, and 83\% sparsity across six representative neuro-symbolic workloads, respectively, with many sparse computational paths based on low activation or probability mass. This observation motivates our \emph{adaptive DAG pruning} (Sec~\ref{subsec:algo_opt_prune}).


\subsection{Unique Characteristics of Neuro-Symbolic vs LLMs}
\label{subsec:profiling_summary}
In summary, neuro-symbolic workloads exhibit distinct characteristics compared to monolithic LLMs in compute kernels, memory behavior, dataflow, and performance scaling.

\textbf{Compute kernels.} LLMs are dominated by regular, highly parallel tensor operations well suited to GPUs. In contrast, neuro-symbolic workloads comprise heterogeneous symbolic and probabilistic kernels with irregular control flow, low arithmetic intensity, and poor cache locality, leading to low GPU utilization and frequent performance bottlenecks.

\textbf{Memory behavior.} Symbolic and probabilistic kernels are primarily memory-bound, operating over large, sparse, and irregular data structures. Probabilistic reasoning further increases memory pressure through large intermediate state caching, creating challenging trade-offs between latency, bandwidth, and on-chip storage.

\textbf{Dataflow and parallelism.} Neuro-symbolic workloads exhibit dynamic and tightly coupled data dependencies. Symbolic and probabilistic computations often depend on neural outputs or require compilation into LLM-compatible structures, resulting in serialized execution, limited parallelism, and amplified end-to-end latency.


\textbf{Performance scaling.} LLMs scale efficiently across GPUs via optimized data and model parallelism. In contrast, symbolic workloads are difficult to parallelize due to recursive control dependencies, while probabilistic kernels incur substantial inter-node communication, limiting scalability on multi-GPUs.

\subsection{Identified Opportunities for Neuro-Symbolic Optimization}
\label{subsec:profiling_opportunities}
While neuro-symbolic systems show promise, improving their efficiency is critical for real-time and scalable deployment. Guided by the profiling insights above, we introduce \NAME (Fig.~\ref{fig:leading}), an algorithm-hardware co-design framework for accelerated probabilistic logical reasoning in neuro-symbolic AI. Algorithmically, a unified representation with adaptive pruning reduces memory footprint (Sec.~\ref{sec:algo_opt}). In hardware architecture, a flexible architecture and dataflow support various symbolic and probabilistic operations (Sec.~\ref{sec:hardware}). \NAME further provides adaptive scheduling and orchestration of heterogeneous LLM-symbolic agentic workloads through a programmable interface (Sec.~\ref{sec:system}). Across reasoning tasks, \NAME consistently boosts performance, efficiency, and accuracy (Sec.~\ref{sec:eval}).

\section{\NAMEnospace: Algorithm Optimizations}
\label{sec:algo_opt}

This section introduces the algorithmic optimizations in \NAME for symbolic and probabilistic reasoning kernels. We present a unified DAG-based computational representation (Sec.~\ref{subsec:algo_opt_dag}), followed by adaptive pruning (Sec.~\ref{subsec:algo_opt_prune}) and regularization techniques (Sec.~\ref{subsec:algo_opt_regularize}) that jointly reduce model complexity and enable efficient neuro-symbolic systems.

\subsection{Stage 1: DAG Representation Unification}
\label{subsec:algo_opt_dag}
\textbf{Motivation.} Despite addressing different reasoning goals, symbolic and probabilistic reasoning kernels often share common underlying computational patterns. For instance, logical deduction in FOL, constraint propagation in SAT, and marginal inference in PCs all rely on iterative graph-based computations. 
Capturing this shared structure is essential to system acceleration. 
DAGs provide a natural abstraction to unify these diverse kernels under a flexible computational model.



\begin{figure}
\begin{minipage}[t!]{\linewidth}
    \centering
    \includegraphics[width=\columnwidth]{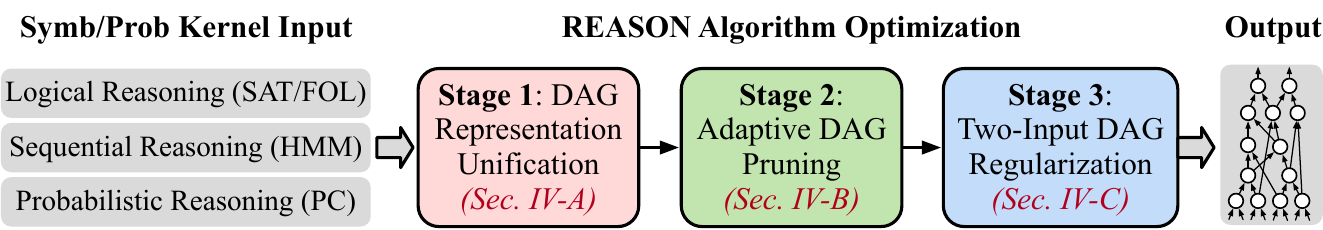}
\end{minipage}%
 \vspace{0.1in}
\begin{minipage}[b]{\columnwidth}
\renewcommand*{\arraystretch}{1.1}
\setlength\tabcolsep{3pt}
\resizebox{\linewidth}{!}{%
\begin{tabular}{l|c|c|c}
\hline
\textbf{Kernel} &
\textbf{DAG Nodes} &
\textbf{DAG Edges} &
\textbf{Inference as DAG Execution} \\
\hline
SAT/FOL &
\begin{tabular}[c]{@{}c@{}}
Literals and\\
logical operators
\end{tabular} &
\begin{tabular}[c]{@{}c@{}}
Logical dependencies between \\
literals, clauses, and formulas
\end{tabular} &
\begin{tabular}[c]{@{}c@{}}
Search and deduction via \\ traversal (DPLL/CDCL)
\end{tabular}
\\ \hline
PC &
\begin{tabular}[c]{@{}c@{}}
Primitive distributions,\\
sum and product nodes
\end{tabular} &
\begin{tabular}[c]{@{}c@{}}
Weighted dependencies encoding\\
probabilistic factorization
\end{tabular} &
\begin{tabular}[c]{@{}c@{}}
Bottom-up probability aggregation\\
and top-down flow propagation
\end{tabular}
\\ \hline
HMM &
\begin{tabular}[c]{@{}c@{}}
Hidden state variables\\
at each time step
\end{tabular} &
\begin{tabular}[c]{@{}c@{}}
State transition and \\
emission dependencies
\end{tabular} &
\begin{tabular}[c]{@{}c@{}}
Sequential message passing\\
(forward–backward, decoding)
\end{tabular}
\\ \hline
\end{tabular}
}
\captionof{figure}{\textbf{Unified DAG representations of neuro-symbolic kernels.} Logical (SAT/FOL), probabilistic (PC), and sequential (HMM) reasoning are expressed using DAG abstraction. Nodes represent atomic reasoning operations, edges encode dependency structure, and graph traversals implement inference procedures. This unification enables shared compilation, pruning, and hardware mapping in \NAMEnospace.}
\label{fig:dag_summary}
\vspace{-10pt}
\end{minipage}
\end{figure}

\textbf{Methodology.} We unify symbolic and probabilistic reasoning kernels under a DAG abstraction, where each node represents an atomic reasoning operation and each directed edge encodes a data/control dependency (Fig.~\ref{fig:dag_summary}). This representation enables a uniform compilation flow -- construction, transformation, and scheduling -- across heterogeneous kernels (logical deduction, constraint solving, probabilistic aggregation, and sequential message passing), and serves as the algorithmic substrate for subsequent pruning and regularization.


\paragraph{For FOL and SAT solvers} DAG nodes represent variables and logical connectives, with edges indicating dependencies between literals and clauses. We represent a propositional CNF formula $\varphi = \bigwedge_{i=1}^m \Bigl(\bigvee_{j=1}^{k_i} l_{ij}\Bigr)$ as DAG with three layers: literal nodes for each literal $l_{ij}$, clause nodes implementing disjunction over literals in $\bigvee_{j} l_{ij}$, and formula nodes implementing conjunction over clauses $\bigwedge_{i}$. In SAT, DAG captures the branching and conflict resolution structures in DPLL/CDCL procedures. In FOL, formulas are encoded as DAGs where inference rules act as graph transformation operators that derive contradictions through node and edge expansion. The compiler converts FOL and SAT inputs (clauses in CNF or quantifier-free predicates) into DAGs via: \emph{Step-\circled{1} Normalization}: predicates are transformed to CNF, removing quantifiers and forming disjunctions of literals. \emph{Step-\circled{2} Node creation}: each literal becomes a leaf node, each clause an OR node over its literals, and the formula an AND node over clauses. \emph{Step-\circled{3} Edge encoding}: edges capture dependencies (literal$\rightarrow$clause$\rightarrow$formula), while watch-lists as metadata.

\paragraph{For PCs} DAG nodes correspond to sum (mixture) or product (factorization) operations $p_n(x)$ over input $x$ (to variable $\mathbf{X}$), with children $ch(x)$. Leaves represent primitive distributions $f_n(x)$. Edges model conditional dependencies. The DAG structure facilitates efficient inference through bottom-up probability evaluation, exploiting structural independence and enabling effective pruning and memorization during probability queries (Eq.~\ref{eq:pc_definition}).
The compiler converts PC into DAGs through: \emph{Step-\circled{1} Graph extraction}: nodes represent random variables, factors, or sub-circuits parsed from expressions such as $P_n(x)$. \emph{Step-\circled{2} Node typing}: arithmetic operators map to sum nodes for marginalization and product nodes for factor conjunction, while leaf nodes store constants or probabilities.



\paragraph{For HMMs}  
The unrolled DAG spans time steps, with nodes representing transition factors $p(z_t | z_{t-1})$ and emission factors $p(x_t|z_t)$ (Eq.~\ref{eq:hmm}), and edges connecting factors across adjacent time steps to reflect Markov dependency. Sequential inference (filtering/smoothing/decoding) becomes structured message passing on this DAG: each step aggregates contributions from predecessor states through transition factors and then applies emission factors.
The compiler converts HMMs into DAGs through: \emph{Step-\circled{1} Sequence unroll}: Each time step becomes a DAG layer, representing states and transitions. \emph{Step-\circled{2} Node mapping}: Product nodes combine transition and emission probabilities; sum nodes aggregate over prior states.

The unified DAG abstraction lays the algorithmic foundation for subsequent pruning, regularization, and hardware mapping, supporting efficient acceleration of neuro-symbolic workloads.

\begin{figure*}[t!]
    \centering
    \includegraphics[width=\linewidth]{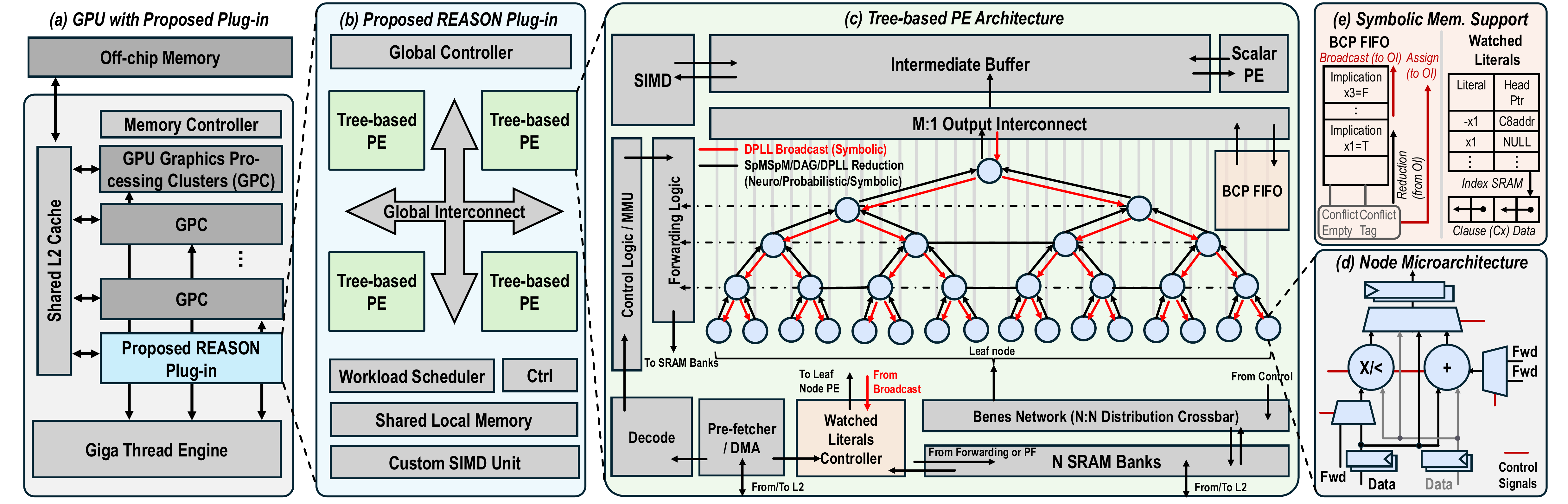}
    \caption{\textbf{Overview of the \NAME hardware acceleration system.} \textbf{(a)} Integration of REASON as a GPU co-processor. \textbf{(b)} \NAME plug-in architecture with PEs, shared local memory, and global scheduling. \textbf{(c)} Tree-based PE architecture enabling broadcast, reduction, and irregular DAG execution. \textbf{(d)} Micro-architecture of a tree node supporting arithmetic and logical operations. \textbf{(e)} FIFO and memory layout supporting symbolic reasoning.}
    \label{fig:arch_top}
    \vspace{-10pt}
\end{figure*}

\subsection{Stage 2: Adaptive DAG Pruning}
\label{subsec:algo_opt_prune}
\textbf{Motivation.} While the unified DAG representation provides a common abstraction, it may contain significant redundancy, such as logically implied literals, inactive substructures, or low-probability paths, that inflate DAG size and degrade performance without improving inference quality.

\textbf{Methodology.} We propose adaptive DAG pruning, a semantics-preserving optimization that identifies and removes redundant paths in symbolic and probabilistic DAGs. For symbolic kernels, pruning targets literals and clauses that are logically redundant. For probabilistic kernels, pruning eliminates low-activation edges that minimally impact inference. This process significantly reduces model size and computational complexity while preserving correctness of logical and probabilistic inference.

\paragraph{Pruning of FOL and SAT via implication graph} 
For SAT solvers and FOL reasoning, we prune redundant literals using implication graphs. Given a CNF formula $\varphi = \bigwedge_i \left(\bigvee_j l_{ij}\right)$, each binary clause $(l \lor l')$ induces two directed implication edges: $\bar{l} \rightarrow l'$ and $\bar{l'} \rightarrow l$. The resulting implication graph captures logical dependencies among literals.
We perform a depth-first traversal to compute reachability relationships between literals. If a literal $l'$ is always implied by another literal $l$, then $l'$ is a hidden literal. Clauses containing both $l$ and $l'$ can safely drop $l'$, reducing clause width without semantic changes. For instance, a clause $C = (l \lor l')$ is reduced to $C' = (l)$. This procedure removes redundant literals (e.g., hidden tautologies and failed literals), preserves satisfiability, and runs in time linear in the size of the implication graph.

\paragraph{Pruning of PCs and HMMs via circuit flow} For probabilistic DAGs such as PCs and HMMs, we prune edges based on probability flow, which quantifies each edge’s contribution to the overall likelihood.

In HMMs, the DAG is unrolled over time steps, with nodes representing transition factors $p(z_t \mid z_{t-1})$ and emission factors $p(x_t \mid z_t)$. We compute expected transition and emission usage via the forward-backward algorithm, yielding posterior state and transition probabilities. Edges corresponding to transitions or emissions with consistently low posterior probability are pruned, as their contribution to the joint likelihood $p(z_{1:T}, x_{1:T})$ is negligible. This pruning preserves inference fidelity while reducing state-transition complexity.


In PCs, sum node $n$ computes $p_n(x) = \sum_{c \in \mathrm{ch}(n)} \theta_{n,c}\, p_c(x)$, where $\theta_{n,c} \ge 0$ denotes the weight associated with child $c$. For an input $x$, we define the circuit flow through edge $(n,c)$ as $F_{n,c}(x) = \frac{\theta_{n,c}\, p_c(x)}{p_n(x)} \cdot F_n(x)$, where $F_n(x)$ denotes the top-down flow reaching node $n$. Intuitively, $F_{n,c}(x)$ measures the fraction of probability mass passing through edge $(n,c)$ for input $x$.
Given a dataset $\mathcal{D}$, the cumulative flow for edge $(n,c)$ is $F_{n,c}(\mathcal{D}) = \sum_{x \in \mathcal{D}} F_{n,c}(x)$.
Edges with the smallest cumulative flow are pruned, as they contribute least to the overall model likelihood. The resulting decrease in average log-likelihood is bounded by $\Delta \log \mathcal{L} \le \frac{1}{|\mathcal{D}|} \sum_{x \in \mathcal{D}} F_{n,c}(x)$, providing a theoretically grounded criterion for safe pruning.



\subsection{Stage 3: Two-Input DAG Regularization}
\label{subsec:algo_opt_regularize}
\textbf{Methodology.} After pruning, the resulting DAGs may still have high fan-in or irregular branching, which hinders efficient hardware execution. To address this, we apply a regularization step that transforms DAGs into a canonical two-input form. Specifically, nodes with more than two inputs are recursively decomposed into balanced binary trees composed of two-input intermediate nodes, preserving the original computation semantics. This normalization promotes uniformity, enabling efficient parallel scheduling, pipelining, and mapping onto \NAME architecture, without sacrificing model fidelity or expressive power.

For each symbolic or probabilistic kernel, the compiler generates an initial DAG, applies adaptive pruning, and then performs two-input regularization to produce a unified balanced representation. These DAGs are constructed offline and used to generate an execution binary that is programmed onto \NAME hardware. This unification-pruning-regularization flow decouples algorithmic complexity from runtime execution and enables predictable performance.


\section{\NAMEnospace: Hardware Architecture}
\label{sec:hardware}
\NAME features flexible co-processor plug-in architecture (Sec.~\ref{subsec:hw_overview}), reconfigurable symbolic/probabilistic PEs (Sec.~\ref{subsec:hw_pe}), flexible support for symbolic and probabilistic kernels (Sec.~\ref{subsec:irregular}-\ref{subsec:hw_symbolic}). Sec.~\ref{subsec:hw_pipeline} presents cycle-by-cycle execution pipeline analysis. Sec.~\ref{subsec:hw_dse} discusses design space exploration and scalability.

\subsection{Overall Architecture}
\label{subsec:hw_overview}
Neuro-symbolic workloads exhibit heterogeneous compute and memory patterns with diverse sparsity, diverging from the GEMM-centric design of conventional hardware. 
Built on the unified DAG representation and optimizations (Fig.~\ref{sec:algo_opt}), \NAME is a reconfigurable and flexible architecture designed to efficiently execute the irregular computations of symbolic and probabilistic reasoning stages in neuro-symbolic AI.

\textbf{Overview.}
\NAME operates as a programmable co-processor tightly integrated with GPU SMs, forming a heterogeneous system architecture. (Fig.~\ref{fig:arch_top}(a)). In this system, \NAME serves as an efficient and reconfigurable ``slow thinking'' engine, accelerating symbolic and probabilistic kernels that are poorly suited to GPU execution.
As illustrated in Fig.~\ref{fig:arch_top}(b), \NAME comprises an array of tree-based PE cores that act as the primary computation engines. A global controller and workload scheduler manage the workload mapping. A shared local memory serves as a unified scratchpad for all cores. Communication between cores and shared memory is handled by a high-bandwidth global interconnect.

\textbf{Tree architecture.} Each PE core is organized as a tree-structured compute engine, as shown in Fig.~\ref{fig:arch_top}(c). Each tree node integrates a specialized datapath, memory subsystem, and control logic optimized for executing DAG-based symbolic and probabilistic operations.

\textbf{Reconfigurable tree engine (RTE).} At the core of each PE is a Reconfigurable Tree Engine (RTE), whose datapath forms a bidirectional tree of PEs (Fig.~\ref{fig:arch_top}(d)). The RTE supports both SAT-style symbolic broadcast patterns and probabilistic aggregation operations. A Benes network interconnect enables N-to-N routing, decoupling SRAM banking from DAG mapping and simplifying compilation of irregular graph structures (Sec.~\ref{subsec:irregular}). Forwarding logic routes intermediate and irregular outputs back to SRAM for subsequent batches.

\textbf{Memory subsystem.} To tackle the memory-bound nature of symbolic and probabilistic kernels, the RTE is backed by a set of dual-port, wide-bitline SRAM banks arranged as a banked L1 cache. A memory front-end with a prefetcher and high-throughput DMA engine moves data from shared scratchpad. A control/memory management unit (MMU) block handles address translation across the distributed memory system.

\textbf{Core control and execution.} A scalar PE acts as the core-level controller, fetching and issuing VLIW-like instructions that configure the RTE, memory subsystem, and specialized units. Outputs from the RTE are buffered before being consumed by the scalar PE or the SIMD Unit, which provides support for executing parallelable subset of symbolic solvers.

\subsection{Reconfigurable Symbolic/Probabilistic PE}
\label{subsec:hw_pe}
The PE architecture is designed to support a wide range of symbolic and probabilistic computation patterns via a VLIW-driven cycle-reconfigurable datapath. Each PE can switch among three operational modes to efficiently execute heterogeneous kernels mapped from the unified DAG representation.

\textbf{Probabilistic mode.}
In probabilistic mode, the node executes irregular DAGs derived from unified probabilistic representations (Sec.~\ref{subsec:irregular}). The nodes are programmed by the VLIW instruction stream to perform arithmetic operations, either addition or multiplication, required by the DAG node mapped onto them. This mode supports probabilistic aggregation patterns such as sum-product computation and likelihood propagation, enabling efficient execution of PCs and HMMs.

\begin{figure}[t!]
    \centering
    \includegraphics[width=\linewidth]{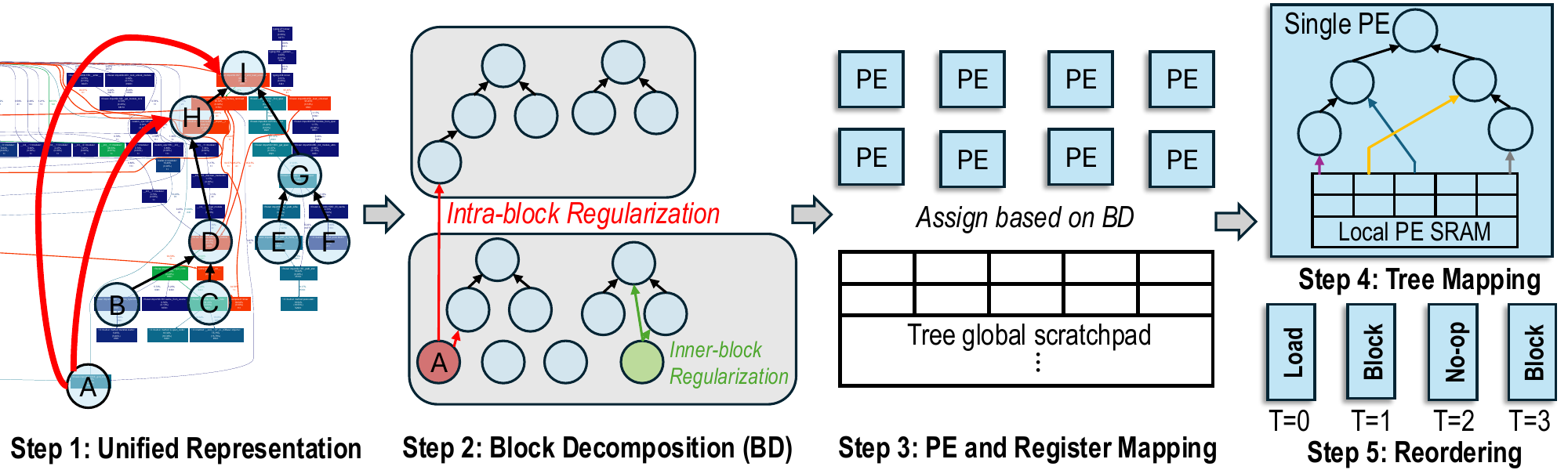}
    \caption{\textbf{Compiler-architecture co-design for probabilistic execution.} A probabilistic DAG is decomposed, regularized, mapped onto tree-based PEs, and scheduled with pipeline awareness to enable efficient execution of irregular probabilistic kernels in \NAMEnospace.}
    \label{fig:arch_pc}
    \vspace{-10pt}
\end{figure}

\textbf{Symbolic mode.}
In symbolic mode, the datapath is repurposed for logical reasoning operations (Sec.~\ref{subsec:hw_symbolic}). Key hardware components are utilized as follows: 
(a) The comparator checks logical states for Boolean Constraint Propagation (BCP), identifying literals as \texttt{TRUE}, \texttt{FALSE}, or \texttt{UNASSIGNED}. 
(b) The adder performs two key functions: address computation by adding the \texttt{Clause Base Address} and \texttt{Literal Index} to locate the next literal in a clause; and clause evaluation by acting as counter to track the number of \texttt{FALSE} literals. This enables fast detection of unit clauses and conflicts, accelerating SAT-style symbolic reasoning.

\textbf{SpMSpM mode.}
The tree-structured PE inherently supports the sparse matrix-matrix multiplication (SpMSpM), a computation pattern widely studied in prior works~\cite{kwon2018maeri,munoz2023flexagon}. In this mode, the leaf nodes are configured as multipliers to compute partial products of the input matrix elements, while the internal nodes are configured as adders to perform hierarchical reductions. This execution pattern allows small-scale neural or neural-symbolic models to be efficiently mapped onto the \NAME engine, extending its applicability beyond purely symbolic and probabilistic kernels.

\subsection{Architectural Support for Probabilistic Reasoning}
\label{subsec:irregular}

Probabilistic reasoning kernels are expressed as DAGs composed of arithmetic nodes (sum and product) connected by data-dependent edges. \NAME exploits its pipelined, tree-structured datapath to efficiently map these DAGs onto parallel PEs. Key architectural features include: multi-tree PE mapping for arithmetic DAG execution, a banked register file with flexible crossbar interconnect to support irregular memory access, and compiler-assisted pipeline scheduling with automatic register management to reduce control overhead. Fig.~\ref{fig:arch_pc} illustrates the overall workflow.

\textbf{Datapath and pipelined execution.} The datapath operates in a pipelined fashion, with each layer of nodes serving as pipeline stages. Each pipeline stage receives inputs from a banked register file, which consists of multiple parallel register banks. Each bank operates independently, providing flexible addressing that accommodates the irregular memory access patterns typical in probabilistic workloads (e.g., PCs, HMMs).

\textbf{Flexible interconnect.} To handle the irregularity in probabilistic DAGs, \NAME employs an optimized interconnection. An input Benes crossbar connects the register file banks to inputs of PE trees, allowing flexible and conflict-free routing of operands into computation units. Output connections from PE to register banks are structured as one-bank-one-PE to minimize hardware complexity while preserving flexibility, balancing trade-offs between utilization and performance.

\begin{figure}[t!]
    \centering
    \includegraphics[width=\linewidth]{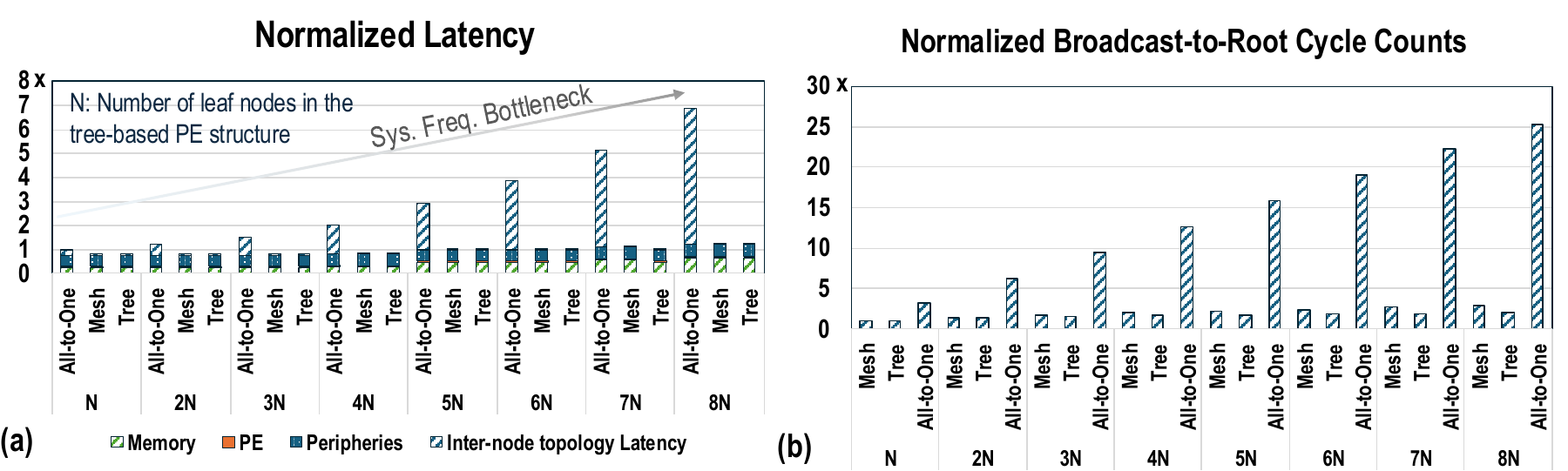}
    \caption{\textbf{Scalability analysis of interconnect topologies.} \textbf{(a)} Normalized latency breakdown as the number of leaf nodes $N$ increases. \textbf{(b)} Normalized broadcast-to-root cycle counts for different PE interconnect structures.}
    \label{fig:physical_driven_result}
    \vspace{-10pt}
\end{figure}

\textbf{Register management.} \NAME adopts an automatic write-address generation policy. Data is written to the lowest available register address in each bank, eliminating the need to encode explicit write addresses in instructions. The compiler precisely predicts these write addresses at compile time due to the deterministic execution sequence, further reducing instruction size and energy overhead.


\textbf{Compiler-driven optimization.} To efficiently translate unified DAGs into executable kernels and map onto hardware datapath, REASON adopts a four-step compiler pipeline (Fig.~\ref{fig:arch_pc}). 

\emph{Step-\circled{1} Block decomposition}: The compiler decomposes the unified DAG from Sec.~\ref{sec:algo_opt} into execution blocks through a greedy search that identifies schedulable subgraphs whose maximum depth does not exceed the hardware tree depth. This process maximizes PE utilization while minimizing inter-block dependencies that may cause read-after-write stalls. The resulting tree-like blocks form the basis for efficient mapping.

\emph{Step-\circled{2} PE mapping}: For each block, the compiler jointly assigns nodes to PEs and operands to register banks, considering topological constraints and datapath connectivity. Nodes are mapped to preserve order, while operands are allocated to banks to avoid simultaneous conflicts. The compiler dynamically updates feasible mappings and prioritizes nodes with the fewest valid options. This conflict-aware strategy minimizes bank contention and balances data traffic across banks.

\emph{Step-\circled{3} Tree mapping}: Once block and register mappings are fixed, the compiler constructs physical compute trees that maximize data reuse in the \NAME datapath. Node fusion and selective replication enhance operand locality and reduce inter-block communication, allowing intermediate results to be consumed within the datapath and lowering memory traffic.

\emph{Step-\circled{4} Reordering}: The compiler then schedules instructions with awareness of the multi-stage pipeline. Dependent operations are spaced by at least one full pipeline interval, while independent ones are interleaved. Lightweight load, store, and copy operations fill idle cycles without disturbing dependencies. Live-range analysis identifies register pressure and inserts minimal spill and reload instructions when needed.

The DAG-to-hardware mapping is an automated heuristic process to generate a compact VLIW program for REASON. Designers can interact for design-space exploration to tune architectural parameters within flexible hardware template.



\subsection{Architectural Support for Symbolic Logical Reasoning}
\label{subsec:hw_symbolic}
To efficiently support symbolic logical reasoning kernels, \NAME features a linked-list-based memory layout and hardware-managed BCP FIFO mechanism (Fig.~\ref{fig:arch_top}(e)), enabling efficient and scalable support to large-scale solver kernels that are fundamental to logical reasoning.

\textbf{Watched literals (WLs) unit.} The WLs unit acts as a distributed memory controller tightly integrated with $N$ SRAM banks, implementing the two-watched-literals indexing scheme in hardware. This design transforms the primary bottleneck in BCP from a sequential scan over the clause database into a selective parallel memory access problem. Crucially, it enables scalability to industrial-scale SAT problems~\cite{moskewicz2001chaff}, where only a small subset of clauses (those on the watch list) need to be accessed at any time. This design naturally aligns with a hierarchical memory system, allowing most clauses to reside in remote scratchpad memory or DRAM, with on-chip SRAM caching only the required clauses indexed by WLs unit.


\begin{figure}[t!]
    \centering
    \includegraphics[width=\linewidth]{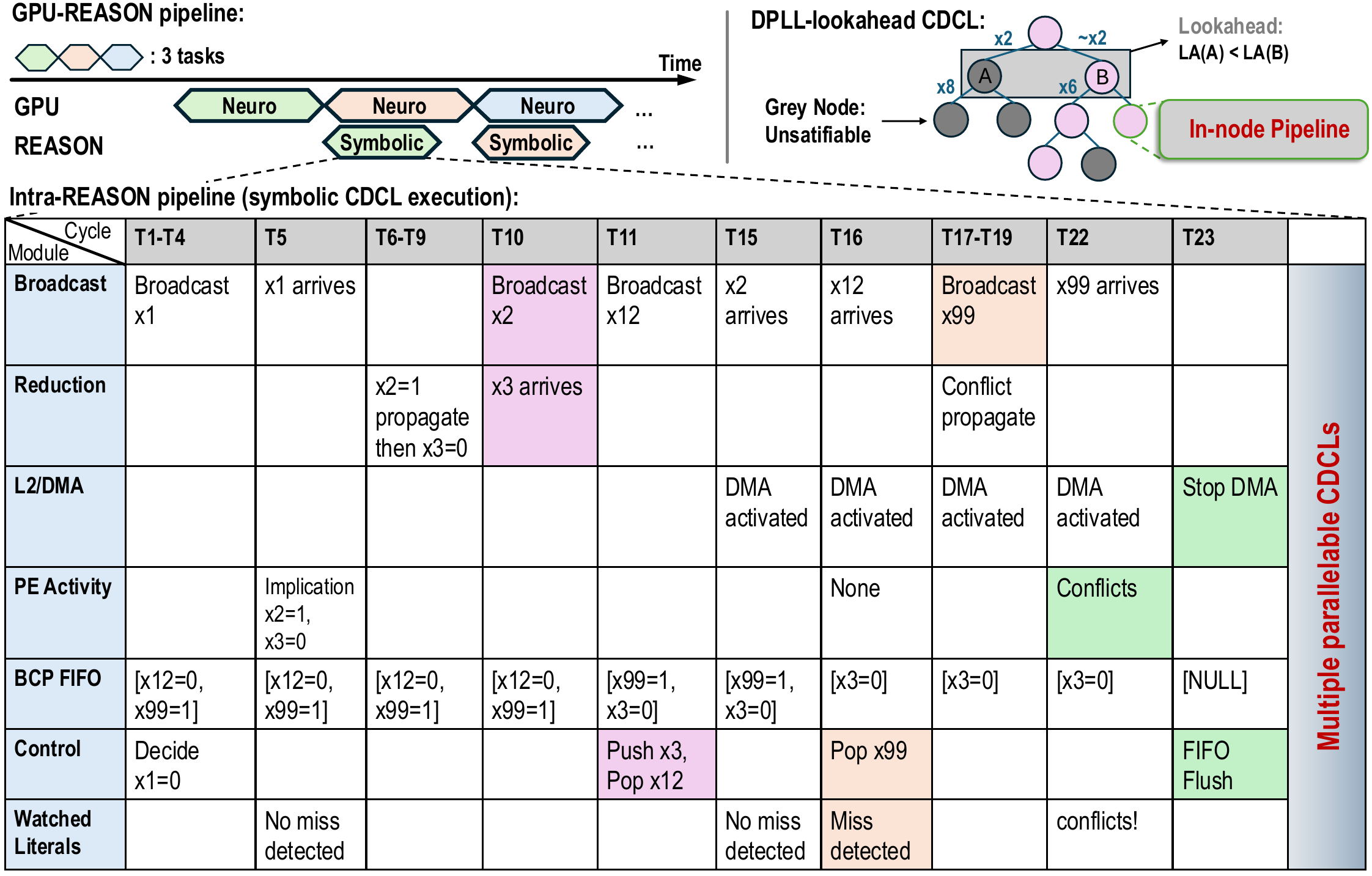}
    \caption{\textbf{Two-level execution pipeline for symbolic reasoning.} Top: task-level overlap between GPU neural execution and REASON symbolic execution. Bottom: detailed cycle-by-cycle timeline of CDCL SAT solving, illustrating pipelined broadcast/reduction, WLs traversal, latency hiding, and priority-based conflict handling. Color represents the causality of hardware events.}
    \vspace{-10pt}
    \label{fig:dpll_cycle}
\end{figure}

\textbf{SRAM layout.}
The local SRAM is partitioned to support a linked-list-based organization of watch lists. A dedicated region stores a head pointer table indexed by literal IDs, each pointing to the start of a watch list, enabling $\mathcal{O}(1)$  access. The main data region stores clauses, each augmented with a next-watch pointer that links to other clauses watching the same literal, forming multiple linked lists within the linear address space. Upon a variable assignment, the WLs unit uses the literal ID to fetch the head pointer and traverses the list by following next-watch pointers, dispatching only the relevant clause to PEs. This hardware-managed indexing eliminates full-database scans and maps efficiently to the adder datapath.

\textbf{BCP FIFO.} The BCP FIFO sits atop the M:1 output interconnect (Fig.~\ref{fig:arch_top}(c)) and serializes multiple parallel implications generated by the leaf tree-node in a single cycle. While many implications can be discovered concurrently, BCP must propagate them sequentially to preserve the causality chain for conflict analysis. The controller immediately broadcasts one implication back into the pipeline, while the rest are queued in the FIFO and processed in a pipelined manner. Within a symbolic (DPLL) tree node, implications are causally independent and can be pipelined, but between assignments, sequential ordering is enforced to maintain correctness. Sec.~\ref{subsec:hw_pipeline} illustrates a detailed cycle-level execution example.


\textbf{Scalability advantages.}
A key advantage of the \NAME architecture is that its tree-based inter-node topology does not become a bottleneck as the symbolic DPLL tree grows (Fig.~\ref{fig:physical_driven_result}(a)). In contrast, all-to-one (or one-to-all) bus interconnects often fail to scale due to post-layout electrical constraints, including high fan-out and buffer insertion for hold-time fixes.
Moreover, given that broadcasting is a dominant operation, the root-to-leaf traversal latency is critical. \NAMEnospace's tree-based inter-node topology achieves exceptional scalability with an $\mathcal{O}(\log N)$ traversal latency, compared to $\mathcal{O}(\sqrt{N})$ for mesh-based designs and $\mathcal{O}(N)$ for bus-based interconnects (Fig.~\ref{fig:physical_driven_result}(b)). This property enables robust scalability for large symbolic reasoning workloads.

\lstset{
    caption=C++ programming interface of {\name}.,
    language=C++,
    frame=single,
    label=listing:api,
    basicstyle=\footnotesize\ttfamily,
    columns=flexible,
    breaklines=true,
    tabsize=4,
    commentstyle=\color{gray}\ttfamily,
    moredelim=**[is][\color{olive}]{@olive@}{@endolive@}, 
}
\begin{figure}
\vspace{-8pt}
\begin{lstlisting}[caption={C++ Programming Interface of \NAME}, label={lst:system_opt_programming_model}]
// Trigger symbolic execution for a single inference 
void REASON_execute(
    int batch_id,  // batch identifier
    int batch_size, // number of objects in the batch
    const void* neural_buffer, // neural results in shared memory
    const void* reasoning_mode, // mode selection
    void* symbolic_buffer // write-back symb. results
    );
// Query current REASON status for a given object
int REASON_check_status (
    int batch_id,   // batch identifier
    bool blocking    // wait till REASON is idle
    );
\end{lstlisting}
\vspace{-10pt}
\end{figure}

\subsection{Case Study: A Working Example of Symbolic Execution}
\label{subsec:hw_pipeline}

Fig.~\ref{fig:dpll_cycle} illustrates the dynamic, pipelined per-node execution of \NAME during a cube-and-conquer SAT solving phase, which highlights several key hardware mechanisms, including inter-node pipelined broadcast/reduction, latency hiding via parallel WLs traversal, and priority-based conflict handling.

Execution begins with the controller issuing a \texttt{Decision} to assign \( x_1 \), which is broadcast through the distribution tree (T1--T4). At T5, leaf nodes concurrently discover two implications: \(x_2\)=\(1\) and \(x_3\)=\(0\). These implications are returned to the controller via the reduction tree in a pipelined manner, where \(x_2\)=\(1\) arrives first, followed by \(x_3\)=\(0\) at T10. Since the FIFO is occupied, \( x_3\)=\(0\) is queued into the BCP FIFO at T11.

At T15, the FIFO pops a subsequent implication (\( x_{12} \)), which triggers WLs lookup. A local SRAM miss prompts the L2/DMA to begin fetching clause, meanwhile BCP FIFO continues servicing queued implications: \( x_{99} \) is popped and broadcast from T17--T19 while DMA fetch is still in progress.


At T22, the propagation of \( x_{99} \) results in a \texttt{Conflict}, which immediately propagates up the reduction tree. Upon receiving the conflict signal at T23, the controller asserts priority control: it halts the ongoing DMA fetch, flushes the FIFO, and discards all pending implications (including \(x_3\)=\(0\)) from the now-invalid search path. The cube-and-conquer phase terminates, and the parallelized DPLL node is forwarded to the scalar PE for CDCL conflict analysis, as discussed in Sec.~\ref{subsec:background_ops}.


\subsection{Design Space Exploration and Scalability}
\label{subsec:hw_dse}
\textbf{Design space exploration.} To identify the optimal configuration of \NAME architecture, we perform a comprehensive design space exploration. We systematically evaluated different configurations by varying key architectural parameters such as the depth of the tree (D), the number of parallel register banks (B), and the number of registers per bank (R). We evaluate each configuration across latency, energy consumption, and energy-delay product (EDP) on representative workloads. The selected configuration (D=3, B=64, R=32) offers the optimal trade-off between performance and energy efficiency.

\textbf{Scalability and variability support.}
Coupled with reconfigurable array, pipeling scheduling, and memory layout optimizations, \NAME provides flexible hardware support across symbolic and probabilistic kernels (e.g., SAT, FOL, PC, HMM), neuro-symbolic workloads, and cognitive tasks, enabling efficient neuro-symbolic processing at scale (Sec.~\ref{sec:eval}).

\textbf{Design choice discussion.} We adopt a unified architecture for symbolic and probabilistic reasoning to maximize flexibility and efficiency, rather than decoupling them into separate engines. We identify these kernels share common DAG patterns, enabling \NAME to execute them on Tree-PEs through a unified representation. This approach achieves $>$90\% utilization with 58\% lower area/power than decoupled designs, while maintaining tight symbolic-probabilistic coupling. A flexible Benes network and compiler co-design handle routing and memory scheduling, ensuring efficient execution.





\section{\NAMEnospace: System Integration and Pipeline}
\label{sec:system}
This section presents the system-level integration of \NAMEnospace. We first present the integration principles and workload partitioning strategy between GPU and \NAME (Sec.~\ref{subsec:system_opt_integration}), then introduce the programming model that enables flexible invocation and synchronization (Sec.~\ref{subsec:system_opt_programming_model}). Finally, we describe the two-level execution pipeline (Sec.~\ref{subsec:system_opt_scheduling}).

\subsection{Integration with GPUs for End-to-End Reasoning}
\label{subsec:system_opt_integration}
\textbf{Integration principles.} As illustrated in Fig.~\ref{fig:arch_top}(a), the proposed \NAME is integrated as a co-processor within GPU system to support efficient end-to-end symbolic and probabilistic reasoning. This integration follows two principles: (1) \underline{\textit{Versatility}} to ensure compatibility with a broad range of logical and probabilistic reasoning workloads in neuro-symbolic pipelines, and (2) \underline{\textit{Efficiency}} to achieve low-latency execution for real-time reasoning.
These principles necessitate careful workload assignment between GPU and \NAME with pipelined execution.

\textbf{Workload partitioning.} To maximize performance while preserving compatibility with existing and emerging LLM-based neuro-symbolic agentic pipelines, we assign neural computation (e.g., LLM inference) to the GPU and offload symbolic reasoning and probabilistic inference to \NAMEnospace. This partitioning exploits the GPU’s high throughput and programmability for neural kernels, while leveraging \NAMEnospace’s reconfigurable architecture optimized for logical and probabilistic operations. It also minimizes data movement and enables pipelined execution: while \NAME processes symbolic reasoning for the current batch, the GPU concurrently executes neural computation for the next batch.

\subsection{Programming Model}
\label{subsec:system_opt_programming_model}
\NAMEnospace's programming model (Listing.~\ref{lst:system_opt_programming_model}) is designed to offer full flexibility and control, making it easy to utilize \NAME for accelerating various neuro-symbolic applications. It exposes two core functions for executing and status checking, enabling acceleration of logical and probabilistic kernels.

\texttt{\NAMEnospace\_execute} is responsible for processing a single symbolic reasoning run. It is called after GPU SMs complete processing the neural LLM. \NAME then performs logical reasoning and probabilistic inference, and writes the symbolic results to shared memory, where SMs use for the next iteration.

\texttt{\NAMEnospace\_check\_status} reports the current execution status of \NAME (\texttt{IDLE} or \texttt{EXECUTION}) and includes an optional blocking flag. This feature allows the host thread to wait for \NAME to complete the current step of reasoning before starting the next, ensuring proper coordination across subtasks without relying on CUDA stream synchronization.

\textbf{Synchronization.} Coordination between SMs and \NAME is handled through shared-memory flag buffers and L2 cache. After executing LLM kernel, SMs write the output to shared memory and set \texttt{neural\_ready} flag. \NAME polls this flag, fetches the data, and performs symbolic reasoning. It then writes the result back to shared memory and sets \texttt{symbolic\_ready} flag, which will be retrieved for final output. This tightly coupled design leverages GPU’s throughput for LLM kernels and \NAMEnospace's efficiency for symbolic reasoning, minimizing overhead and maximizing performance.

\begin{table}[t!]
\scriptsize
\centering
\caption{\textbf{Hardware baselines.} Comparison of device specs.}
\renewcommand*{\arraystretch}{0.95}
\setlength\tabcolsep{3pt}
\resizebox{\linewidth}{!}{%
\begin{threeparttable}
\begin{tabular}{l|c|c|c|c|c}
\hline
\textbf{Device} & \textbf{Technology} & \textbf{SRAM} & \textbf{Number of Cores} & \textbf{Area [$\mathrm{mm}^2$]} & \textbf{Power [W]} \\
\hline
Orin NX~\cite{onx} & 8 nm & 4 MB & 512 CUDA Cores & 450 & 15 \\
RTX A6000~\cite{a6000} & 8 nm & 16.5 MB & 10572 CUDA Cores & 628 & 300 \\
Xeon CPU~\cite{xeoncpu} & 10 nm & 112.5 MB & 60 cores per socket & 1600 & 270 \\
TPU~\cite{jouppi2021ten} & 7 nm & 170 MB & 8 128$\times$128 PEs & 400 & 192 \\
DPU\textsuperscript{\#}~\cite{shah2022dpu} & 28 nm & 2.4 MB &  8 PEs / 56 Nodes & 3.20 & 1.10 \\
\NAME & 28 nm & 1.25 MB & 12 PEs / 80 Nodes & 6.00 & 2.12 \\
\NAMEnospace\textsuperscript{*} & 12 nm & 1.25 MB &  12 PEs / 80 Nodes & 1.37 & 1.21 \\
\NAMEnospace\textsuperscript{*} & 8 nm & 1.25 MB &  12 PEs / 80 Nodes & 0.51 & 0.98 \\
\hline
\end{tabular}
\begin{tablenotes}
        \item[*] \scriptsize The 12nm and 8nm data are scaled from the \textit{DeepScaleTool}~\cite{sarangi2021deepscaletool} with a voltage of 0.8V and a frequency of 500MHz. \textsuperscript{\#} The terminology for tree-based architecture is renamed from \texttt{tree} to \texttt{PE} and \texttt{PE} to \texttt{node} to align with the proposed REASON.
      \end{tablenotes}
    \end{threeparttable}
}
\label{tab:device_specs}
\vspace{-5pt}
\end{table}

\subsection{Two-Level Execution Pipeline}
\label{subsec:system_opt_scheduling}
Our system-level design employs a two-level pipelined execution model (Fig.~\ref{fig:dpll_cycle} top-left) to maximize concurrency across neural and symbolic kernels. The \emph{GPU-REASON pipeline} overlaps the execution of symbolic kernels on REASON for step~$N$ with neural kernels on GPU for step~$N$+1, effectively hiding the latency of one stage and improving throughput. Within \NAMEnospace, the \emph{Intra-REASON pipeline} exploits inter-node pipelined broadcast and reduction to hide communication latency, using parallel worklist traversal and priority-based conflict handling to accelerate symbolic kernels (Sec.~\ref{subsec:hw_pipeline}). The compiler integrates pipeline-aware scheduling to reorder instructions, avoid read-after-write hazards, and insert no-operation instructions when necessary, ensuring each stage receives valid data without interruption.



\begin{figure}[t!]
    \centering
    \includegraphics[width=\linewidth]{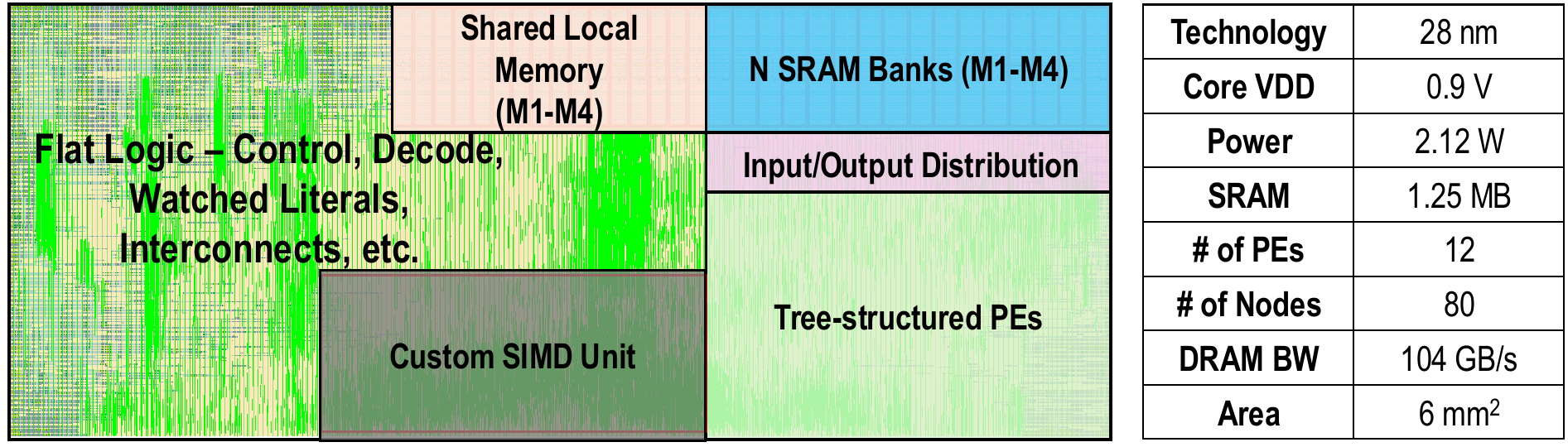}
    \caption{\textbf{REASON layout and specifications.} The physical design and key operating specifications of our proposed \NAME hardware.}
    \label{fig:pnr}
    \vspace{-10pt}
\end{figure}

\section{Evaluation}
\label{sec:eval}

This section introduces \NAME evaluation settings (Sec.~\ref{subsec:eval_setup}), and benchmarks the proposed algorithm optimizations (Sec.~\ref{subsec:eval_algo}) and hardware architecture (Sec.~\ref{subsec:eval_arch}).

\subsection{Evaluation Setup}
\label{subsec:eval_setup}

\textbf{Datasets.} We evaluate \NAME on 10 commonly-used reasoning datasets: IMO~\cite{trinh2024solving}, MiniF2F~\cite{zheng2021minif2f}, TwinSafety~\cite{kang2024r}, XSTest~\cite{rottger2023xstest}, CommonGen~\cite{lin2020commongen}, News~\cite{zhang2020pointer}, CoAuthor~\cite{lee2022coauthor}, AwA2~\cite{xian2018zero}, FOLIO~\cite{han2022folio}, and ProofWriter~\cite{tafjord2020proofwriter}. The tasks are measured by the reasoning and deductive accuracy. 

\textbf{Algorithm setup.} We evaluate \NAME on six state-of-the-art neuro-symbolic models, i.e., AlphaGeometry~\cite{trinh2024solving}, R$^2$-Guard~\cite{kang2024r}, GeLaTo~\cite{zhang2023tractable}, Ctrl-G~\cite{zhang2024adaptable}, NeuroPC~\cite{chen2025neural}, and LINC~\cite{olausson2023linc}. Following the setup in the original literature, we determine the hyperparameters based on end-to-end reasoning performance on the datasets. Our proposed \NAME algorithm optimizations are general and can work as a plug-and-play extension to existing neuro-symbolic algorithms.

\textbf{Baselines.} We consider several hardware baselines, including Orin NX~\cite{onx} (since our goal is to enable real-time neuro-symbolic at edge), RTX GPU~\cite{a6000}, Xeon CPU~\cite{xeoncpu}, ML accelerators (TPU~\cite{jouppi2021ten}, DPU~\cite{shah2022dpu}). Tab.~\ref{tab:device_specs} lists configurations.

\textbf{Hardware setup.} We implement \NAME architecture with ~\cite{shah2022dpu} as the baseline template, synthesize with Synopsys DC~\cite{compiler}, and place and route using Cadence Innovus~\cite{innovus} at TSMC 28nm node. Fig.~\ref{fig:pnr} illustrates the layout and key specifications. The \NAME hardware consumes an area of 6~mm$^2$ and an average power of 2.12~W across workloads based on Synopsys PTPX~\cite{ptpx} power-trace analysis (Fig.~\ref{fig:eval_energy}(a)).
Unlike conventional tree-based arrays that mainly target neural workloads, \NAME provides unified, reconfigurable support for neural, symbolic, and probabilistic computation.

\textbf{Simulation setup.} To evaluate end-to-end performance of \NAME when integrated with GPUs, we develop a cycle-accurate simulator based on Accel-Sim (built on GPGPU-Sim)~\cite{khairy2020accel}. 
The simulator is configured for Orin NX architecture. The on-chip GPU is modeled with 8~SMs, each supporting 32 threads per warp, 48~KB shared memory, and 128~KB L1 cache, with a unified 2~MB L2 cache shared across SMs. The off-chip memory uses a 128-bit LPDDR5 interface with 104~GB/s peak BW. DRAM latency and energy are modeled using LPDDR5 timing parameters.

\textbf{Simulator test trace derivation.} We use GPGPU-Sim to model interactions between SMs and \NAMEnospace, including transferring neural results from SMs to \NAME and returning symbolic reasoning results from \NAME to SMs. To simulate communication overhead, we extract memory access traces from neuro-symbolic model execution on Orin, capturing data volume and access patterns as inputs to GPGPU-Sim for accurate modeling. For GPU comparison baselines, we use real hardware measurements to get accurate ground-truth data.

\subsection{\NAME Algorithm Performance}
\label{subsec:eval_algo}

\begin{table}[t!]
\scriptsize
\centering
\caption{\textbf{\NAME algorithm optimization performance.} \textnormal{\NAME achieves comparable accuracy with reduced memory footprint via unified DAG representation, adaptive pruning, and regularization.}}
\renewcommand*{\arraystretch}{1}
\setlength\tabcolsep{2.2pt}
\resizebox{\linewidth}{!}{%
\begin{tabular}{c|c|c|c|cc}
\hline
\multirow{2}{*}{\textbf{Workloads}} & \multirow{2}{*}{\textbf{Benchmarks}} & \multirow{2}{*}{\textbf{Metrics}} & \multirow{2}{*}{\textbf{\begin{tabular}[c]{@{}c@{}}Baseline\\ Performance\end{tabular}}} & \multicolumn{2}{c}{\textbf{After \NAME Algo. Opt.}}                    \\ \cline{5-6} 
                                   &                                     &                                  &                                & \multicolumn{1}{c|}{\textbf{Performance}} & \textbf{Memory $\downarrow$} \\ \hline
\multirow{2}{*}{AlphaGeo}             & IMO                                 & Accuracy ($\uparrow$)                        & 83\%                           & \multicolumn{1}{c|}{83\%}         & 25\%                   \\ \cline{2-6} 
                                   & MiniF2F                             & Accuracy ($\uparrow$)                        & 81\%                           & \multicolumn{1}{c|}{81\%}         & 21\%                   \\ \hline
\multirow{2}{*}{R$^2$-Guard}            & TwinSafety                                  & AUPRC ($\uparrow$)                           & 0.758                          & \multicolumn{1}{c|}{0.752}        & 37\%                   \\ \cline{2-6} 
                                   & XSTest                              & AUPRC ($\uparrow$)                           & 0.878                          & \multicolumn{1}{c|}{0.881}        & 30\%                   \\ \hline
\multirow{2}{*}{GeLaTo}            & CommonGen                           & BLEU ($\uparrow$)                       & 30.3                           & \multicolumn{1}{c|}{30.2}         & 41\%                   \\ \cline{2-6} 
                                   & News                                & BLEU ($\uparrow$)                     & 5.4                            & \multicolumn{1}{c|}{5.4}          & 27\%                   \\ \hline
Ctrl-G                             & CoAuthor                            & Success rate ($\uparrow$)                     & 87\%                           & \multicolumn{1}{c|}{86\%}         & 29\%                   \\ \hline
NeuroSP                            & AwA2                                & Accuracy                         & 87\%                        & \multicolumn{1}{c|}{87\%}      & 43\%                   \\ \hline
\multirow{2}{*}{LINC}              & FOLIO                               & Accuracy ($\uparrow$)                     & 92\%                           & \multicolumn{1}{c|}{91\%}         & 38\%                   \\ \cline{2-6} 
                                   & ProofWriter                         & Accuracy ($\uparrow$)                     & 84\%                           & \multicolumn{1}{c|}{84\%}         & 26\%                   \\ \hline
\end{tabular}
}
\label{tab:eval_algo}
\vspace{-3pt}
\end{table}

\textbf{Reasoning accuracy.} To evaluate \NAME algorithm optimization (Sec.~\ref{sec:algo_opt}), we benchmark it on ten reasoning tasks (Sec.~\ref{subsec:eval_setup}). Tab.~\ref{tab:eval_algo} lists the arithmetic performance and DAG size reduction. We observe that \NAME achieves comparable reasoning accuracy through unification and adaptive DAG pruning. Through pruning and regularization, \NAME enables 31.7\% memory footprint savings on average of ten reasoning tasks across six neuro-symbolic workloads.

\subsection{\NAME Architecture Performance}
\label{subsec:eval_arch}

\begin{figure}[t!]
\centering
\includegraphics[width=\columnwidth]{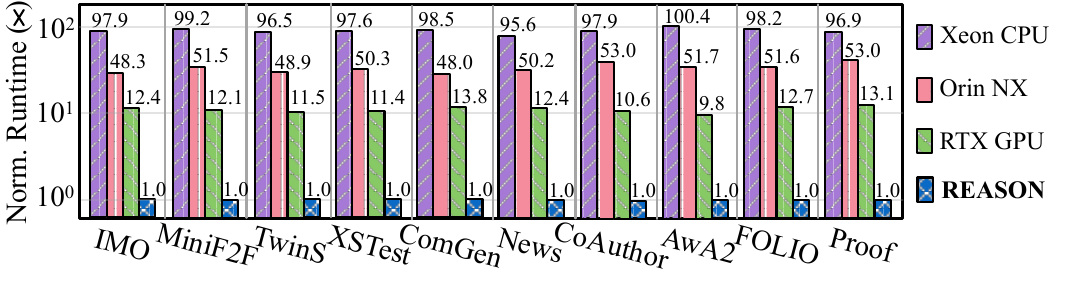}
\vspace{-0.15in}
        \caption{\textbf{End-to-end runtime improvement.} \textnormal{\NAME consistently outperforms Xeon CPU, Orin NX, and RTX GPU in end-to-end runtime latency evaluated on 10 logical and cognitive reasoning tasks.}}
        \label{fig:eval_e2e_runtime}
        \vspace{-10pt}
\end{figure}

\textbf{Performance improvement.} We benchmark \NAME accelerator with Orin NX, RTX GPU, and Xeon CPU for accelerating neuro-symbolic algorithms on 10 reasoning tasks (Fig.~\ref{fig:eval_e2e_runtime}). For GPU baseline, for neuro kernels, we use Pytorch package that leverages CUDA and cuBLAS/cuDNN libraries; for symbolic kernels, we implement custom kernels optimized for logic and probabilistic operations. The workload is tiled by CuDNN in Pytorch based on block sizes that fit well in GPU memory.
We observe that \NAME exhibits consistent speedup across datasets, e.g., 50.65$\times$/11.98$\times$ speedup over Orin NX and RTX GPU. Furthermore, \NAME achieves real-time performance ($<$1.0 s) on solving math and cognitive reasoning tasks, indicating that \NAME enables real-time probabilistic logical reasoning-based neuro-symbolic system with superior reasoning and generalization capability, offering a promising solution for future cognitive applications.

\begin{figure}[t!]
\centering
\includegraphics[width=\columnwidth]{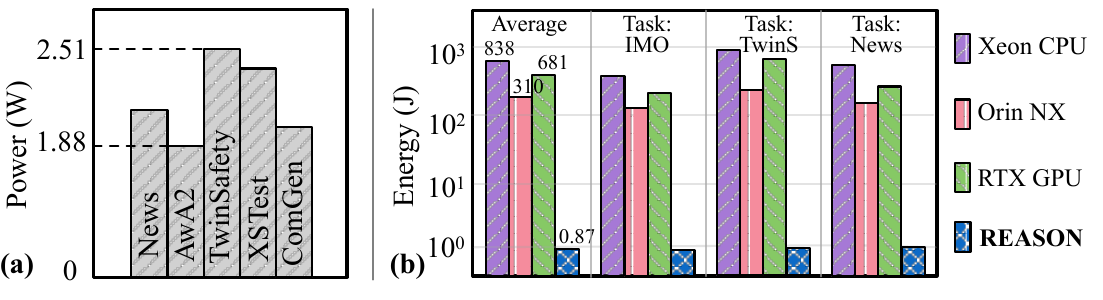}
\vspace{-0.15in}
        \caption{\textbf{Energy efficiency improvement.} \textnormal{\textbf{(a)} Power analysis of \NAME across workloads. \textbf{(b)} Energy efficiency comparison between \NAME and CPUs/GPUs, evaluated from 10 reasoning tasks.}}
        \label{fig:eval_energy}
    \vspace{-10pt}
\end{figure}

\begin{figure*}[t!]
\centering
\vspace{-5pt}
\includegraphics[width=1.9\columnwidth]{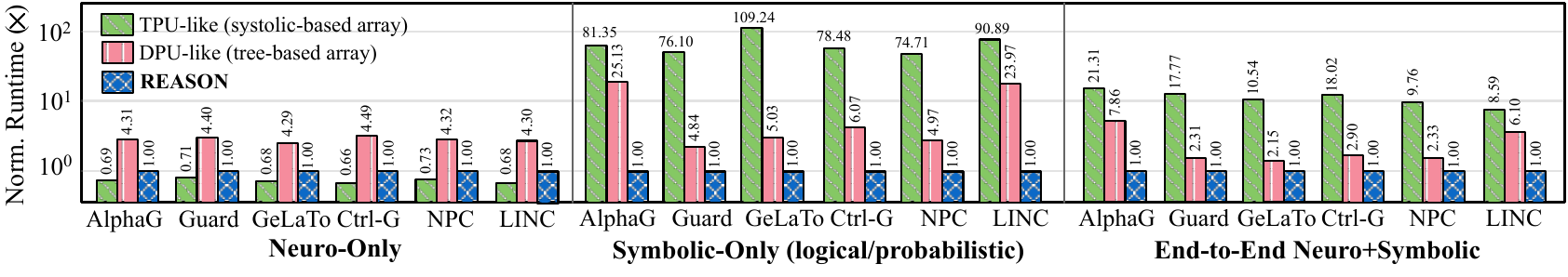}
        \vspace{-0.1in}
        \caption{\textbf{Improved efficiency over ML accelerators.}
        \textnormal{Speedup comparison of neural, symbolic (logical and probabilistic), and end-to-end neuro-symbolic system over TPU-like systolic-based array and DPU-like tree-based array architecture.}}
        \label{fig:ml_acc_comp}
        \vspace{-10pt}
\end{figure*}


\textbf{Energy efficiency improvement.} \NAME accelerator achieves two orders of energy efficiency than Orin NX, RTX GPU, and Xeon CPU consistently across workloads (Fig.~\ref{fig:eval_energy}(b)). To further assess \NAME energy efficiency in long-term deployment, we perform consecutive tests on \NAME using mixed workloads, incorporating both low-activity and high-demand periods, with 15s idle intervals between scenarios. On average, \NAME achieves 681$\times$ energy efficiency compared to RTX GPU. Additionally, when compared to V100 and A100, \NAME shows 4.91$\times$ and 1.60$\times$ speedup, with 802$\times$ and 268$\times$ energy efficiency, respectively.


\textbf{Compare with CPU+GPU.} We compare the performance of RESAON as GPU plug-in over the CPU+GPU architecture across neuro-symbolic workloads. CPU+GPU architecture is not efficient for neuro-symbolic computing due to (1) high latency of symbolic/probabilistic operations with poor locality and $<$5\% CPU parallel efficiency, (2) $>$15\% inter-device communication overhead from frequent neural-symbolic data transfers, and (3) fine-grained coupling between neural and symbolic modules that makes handoffs costly. \NAME co-locates logical and probabilistic reasoning beside GPU SMs, sharing L2 and inter-SM fabric to eliminate transfers and pipeline neural-symbolic execution.


\textbf{Compare with ML accelerators.} We benchmark the runtime of neural and symbolic operations on TPU-like systolic array~\cite{jouppi2021ten} and DPU-like tree-based array~\cite{shah2022dpu} over different neuro-symbolic models and tasks (Fig.~\ref{fig:ml_acc_comp}). 
For TPU-like architecture, we use SCALE-Sim~\cite{raj2025scale}, configured with eight 128$\times$128 systolic arrays.
For DPU-like architecture, we use MAERI~\cite{kwon2018maeri}, configured with eight PEs in 56-node tree-based array.
Compared with ML accelerators, we observe that \NAME achieves similar performance in neural operations, while exhibiting superior symbolic logic and probabilistic operation efficiency thus end-to-end speedup in neuro-symbolic systems. 


\begin{table}[t!]
\vspace{0.05in}
\centering
\caption{\textbf{Ablation study of necessity of co-design.} \textnormal{The normalized runtime achieved by \NAME framework w/o the proposed algorithm optimization or hardware techniques on different tasks.}}
\renewcommand*{\arraystretch}{1.2}
\setlength\tabcolsep{1.6pt}
\resizebox{\linewidth}{!}{%
\begin{tabular}{c|ccccc}
\hline
Neuro-symbolic System & \multicolumn{5}{c}{Normalized Runtime (\%) on} \\
Algorithm @ Hardware & IMO~\cite{trinh2024solving}  & MiniF~\cite{zheng2021minif2f}     & TwinS~\cite{kang2024r} &  XSTest~\cite{rottger2023xstest} & ComG~\cite{lin2020commongen}    \\ \hline
\cite{trinh2024solving,kang2024r,zhang2023tractable} @ Orin NX &  100     &   100    & 100  & 100 & 100    \\ \hline
\textbf{\NAME Algo. @ Orin NX} &  84.2\%     & 87.0\%     &  78.3\%  & 82.9\%	& 86.6\%   \\ \hline
\textbf{\NAME Algo. @ \NAME HW}  &   2.07\%    & 1.94\%     &   2.04\% & 1.99\% & 2.08\%   \\ \hline
\end{tabular}
}
\label{tab:ablation_codesign}
\vspace{-10pt}
\end{table}

\textbf{Ablation study on the proposed hardware techniques.} \NAME features reconfigurable tree-based array architecture, efficient register bank mapping, and adaptive scheduling strategy to reduce compute latency for neural, symbolic, and probabilistic kernels (Sec.~\ref{sec:hardware} and Sec.~\ref{sec:system}). To demonstrate the effectiveness, we measure the runtime of \NAME w/o the scheduling, reconfigurable architecture, and bank mapping. In particular, the proposed memory layout support can trim down the runtime by 22\% on average. Additionally, with the proposed reconfigurable array and scheduling strategy, the runtime reduction ratio can be further enlarged to 56\% and 73\%, indicating that both techniques are necessary for \NAME to achieve the desired efficient reasoning capability.

\textbf{Ablation study of the necessity of co-design.} To verify the necessity of algorithm-hardware co-design strategy to achieve efficient probabilistic logical reasoning-based neuro-symbolic systems, we measure the runtime latency of our \NAME w/o the proposed algorithm or hardware techniques in Tab.~\ref{tab:ablation_codesign}. Specifically, with our proposed \NAME algorithm optimization, we can trim down the runtime to 78.3\% as compared to R$^2$-Guard~\cite{kang2024r} on the same Orin NX hardware and TwinSafety task. Moreover, with both proposed \NAME algorithm optimization and accelerator, the runtime can be reduced to 2.04\%, indicating the necessity of the co-design strategy of \NAME framework.

\textbf{REASON neural optimization.} \NAME accelerates symbolic reasoning and enables seamless interaction with GPU optimized for neuro (NN/LLM) computation. To further optimize neural module, we integrate standard LLM acceleration techniques: memory-efficient attention~\cite{kwon2023efficient}, chunked prefill~\cite{vllm_optimization}, speculative decoding~\cite{leviathan2023fast}, FlashAttention-3 kernels~\cite{shah2024flashattention}, FP8 KV-Cache quantization~\cite{vllm_quant}, and prefix caching~\cite{vllm_caching}. These collectively yield 2.8-3.3$\times$ latency reduction for unique prompts and 4-5$\times$ when prefixes are reused. While REASON’s reported gains stem from its hardware-software co-design, these LLM optimizations are orthogonal and can be applied in conjunction to unlock full potential system speedup.

\section{Related Work}
\label{sec:related_work}

\textbf{Neuro-symbolic AI.}
Neuro-symbolic AI has emerged as a promising paradigm for addressing limitations of purely neural models, including factual errors, limited interpretability, and weak multi-step reasoning.~\cite{zhang2020alphazero,badreddine2022logic,hohenecker2020ontology,dongneural,pryor2022neupsl,manhaeve2021neural,yang2020neurasp}. Systems such as LIPS~\cite{li2025proving}, AlphaGeometry~\cite{trinh2024solving}, NSC~\cite{mao2025neuro}, NS3D~\cite{hsu2023ns3d} demonstrate strong performance across domains ranging from mathematical reasoning to embodied and cognitive robotics.
However, most prior work focuses on algorithmic design and model integration. \emph{\NAME systematically characterizing the architectural and system-level properties of probabilistic logical reasoning in neuro-symbolic AI, and proposes an integrated acceleration framework for scalable deployment.}


\textbf{System and architecture for neuro-symbolic AI.}
Early neuro-symbolic systems largely focused on software-level abstractions, such as training semantics and declarative languages that integrate neural with logical or probabilistic reasoning, such as DeepProbLog~\cite{manhaeve2018deepproblog}, DreamCoder~\cite{ellis2023dreamcoder}, and Scallop~\cite{li2023scallop}. Recent efforts have begun to address system-level challenges, such as heterogeneous mapping, batching control-heavy reasoning, and kernel specialization, including benchmarking~\cite{wan2024towards_ispass}, pruning~\cite{dang2022sparse}, Lobster~\cite{biberstein2025lobster}, Dolphin~\cite{naik2024dolphin}, and KLay~\cite{maene2024klay}. At the architectural level, a growing body of work exposes the mismatch between compositional neuro-symbolic workloads and conventional hardware designs, motivating cognitive architectures such as CogSys~\cite{wan2025cogsys}, NVSA architectures~\cite{wan2024towards}, and NSFlow~\cite{yang2025nsflow}.
\textit{REASON advances this direction with the first flexible system-architecture co-design that supports probabilistic logical reasoning-based neuro-symbolic AI and integrates with GPU, enabling efficient and scalable compositional neuro-symbolic and LLM+tools agentic system deployment.}


\section{Conclusion}
\label{sec:conclusion}
We present \NAMEnospace, the integrated acceleration framework for efficiently executing probabilistic logical reasoning in neuro-symbolic AI. \NAME introduces a unified DAG abstraction with adaptive pruning and a flexible reconfigurable architecture integrated with GPUs to enable efficient end-to-end execution. Results show that system-architecture co-design is critical for making neuro-symbolic reasoning practical at scale, and position \NAME as a potential foundation for future agentic AI and LLM+tools systems that require structured and interpretable reasoning alongside neural computation.
\section*{Acknowledgements}
This work was supported in part by CoCoSys, one of seven centers in JUMP 2.0, a Semiconductor Research Corporation (SRC) program sponsored by DARPA. We thank Ananda Samajdar, Ritik Raj, Anand Raghunathan, Kaushik Roy, Ningyuan Cao, Katie Zhao, Alexey Tumanov, Shirui Zhao, Xiaoxuan Yang, Zhe Zeng, and the anonymous HPCA reviewers for insightful discussions and valuable feedback.


\bibliographystyle{unsrt}
\bibliography{refs}

\end{document}